\documentclass[review]{elsarticle}

\usepackage{hyperref}
\usepackage{amssymb}

\journal{??}

 \biboptions{round,semicolon,sort,authoryear}

\begin{document}

\begin{frontmatter}

\title{The Influence of Feature Representation of Text \\ on the Performance of Document Classification}

\author [odjel] {Sanda Martin\v{c}i\'c-Ip\v{s}i\'c\corref{corauthor}}
\ead{smarti@uniri.hr}
\address [odjel]{Department of Informatics, University of Rijeka; Rijeka, Croatia }
\cortext[corauthor]{Corresponding Author: Department of Informatics, University of Rijeka, Radmile Matej\v{c}i\'c 2, 51000 Rijeka, Croatia, +385 51 584 714}

\author [odjel] {Tanja  Mili\v{c}i\'c}
\ead{tanja.milicic@student.uniri.hr}

\author[FU,IJS] {Ljup\v{c}o Todorovski}
\ead{Todorovski@fu.uni-lj.si}

\address[FU]{University of Ljubljana, Faculty of Administration, Ljubljana, Slovenia}
\address[IJS]{Institute Jo\v{z}ef Stefan, Department of Knowledge Technologies, Ljubljana}






\begin{abstract}
In this paper we perform a comparative analysis of three models for feature representation of text documents in the context of document classification. In particular, we consider the most often used family of models bag-of-words, recently proposed continuous space models word2vec and doc2vec, and the model based on the representation of text documents as language networks. While the bag-of-word models have been extensively used for the document classification task, the performance of the other two models for the same task have not been well understood. This is especially true for the network-based model that have been rarely considered for representation of text documents for classification. In this study, we measure the performance of the document classifiers trained using the method of random forests for features generated the three models and their variants. The results of the empirical comparison show that the commonly used bag-of-words model has performance comparable to the one obtained by the emerging continuous-space model of doc2vec. In particular, the low-dimensional variants of doc2vec generating up to 75 features are among the top-performing document representation models. The results finally point out that doc2vec shows a superior performance in the tasks of classifying large documents.

\end{abstract}


\end{frontmatter}

\section{Introduction}
The growth of the use of electronic documents propelled the development of solutions aiming at automatic organization of those documents in appropriate categories. The related task of automatic classification of text documents become an important tool for the relevant applications of news filtering and organization, information retrieval, opinion mining, spam filtering and e-mail classification~ \citep{aggarwal2012survey}. In general, document classification is the task of assigning a label from a predefined set of candidate class labels to a text document of interest. More formally, the task of a single-label document classification can be defined as follows~\citep{sebastiani2002machine}: Let $D$ be a set of documents, and $C$ a set of class labels. Given a set of training pairs $\langle d_i, c_i \rangle \in D \times C$, we seek to construct a classification model $f : D \rightarrow C$ such that the set of misclassified documents is minimized as much as possible. In turn, the classification model $f$ can be used for predicting the class label of any given document.

Learning a classification model for documents is very similar to the standard supervised machine learning task, where each training example is annotated with the correct class label. However, before applying the standard methods for supervised machine learning to the task of document classification, we have to resolve the issue of representing documents as vectors of feature values. More formally, we need a document representation model $m: D \rightarrow R^n$, where $n$ corresponds to the number of features representing documents. Model $ m $ transforms a given document $ d $ to a $n$-dimensional vector of real-valued features. In the bag-of-words models, each feature corresponds to a word and its value to the frequency of that word in the document. The continuous space representation models (word2vec/doc2vec) embed the words/documents in a normed vector space, where the closeness of vectors corresponds to the word/document semantic similarity: the features correspond to the dimensions of the vector space. In the network-based models, features are structural properties of the document network, which represents a document as a graph with vertices corresponding to words and edges denoting the co-occurrence of words in sentences.

In this paper, we empirically analyze the influence of the document representation models on the performance of document classification. The empirical analysis is performed on seven benchmark tasks of document classification stemming from four standard data sets used in numerous studies~\citep{francis1979brown, lang1995newsweeder, craven1998learning, lewis2004rcv1}. Our primary focus is on identifying the variant of the three document representation models, introduced above, which leads to the best classification performance. Hence, in all the experiments, we use a strong versatile classification model of a random forest~\citep{Breiman2001} and a single dimension-reduction method of principal component analysis~\citep{Jolliffe2014}, where necessary. We analyze the document classification performance from different perspectives corresponding to the standard measures of classification accuracy, recall, precision, F1-score and the area under the receiver operating characteristic curve.

The paper represents an important contribution to the existing work on the comparative analysis of document classification performance. While the performance of different variants of the bag-of-words model is well studied~\citep{Yang1999, sebastiani2002machine, Forman2003, aggarwal2012survey}, the systematic comparative study of the performance of the other two document representation models is missing. Namely, the comparative studies focus on identifying the best performing classification model and/or subset of the bag-of-words features. Moreover, while recent studies of document representation models widely consider the continuous space models of word2vec and doc2vec, a network-based models have not been considered in the machine learning literature and have been applied in the context of document classification sporadically. Therefore, this paper provides the first systematic comparative analysis that include the widely used bag-of-words models, the emerging vector space models, and the network-based models that have been neglected. To sum up, this comparative study will contribute a novel and relevant guide for deciding upon the appropriate document representation model for a given document classification task.

The rest of the paper is organized as follows. In Section~2, we introduce the three document representation models and their variants as well as provide an overview of related studies for each of them. Section~3 introduces the setup used to conduct the empirical comparison of the performance of the document representation models for document classification. Section~4 presents and discusses the experimental results. Finally, Section~5 summarizes the contributions of the paper and outlines the directions for further research.

\section{Document Representation Models}

We can cluster the document representation models into two large groups. The models in the first group lead to features that are at the level of words, while models in the second construct features at the level of the whole document. The bag-of-words model clearly belongs to the first group, where features correspond to words and feature values to word presence/absence or frequency. The word2vec model is a representative of the first group as well, since it relies on the embedding of words in a vector space. On the other hand, the doc2vec model operates at the document level, since it provides the embedding of whole documents in a vector space. Finally, note that the network-based model belongs to both groups. Some of the network-based features that quantify the properties of individual network nodes (recall that these denote words) correspond to the first group, while others quantifying the properties of the entire network (document) correspond to the second group.

In the continuation of this section, we are going to provide a detailed introduction of the three document representation models compared in this study.

\subsection{Bag-of-Words Model}

The bag-of-words (BOW) model represents each document as an unordered set (bag) of features that correspond to the terms in a vocabulary for a given document collection. The vocabulary can include words, a sequence of words (\emph{token n-grams}) or sequences of letters of length $n$ (\emph{character n-grams})~\citep{manning2008h, Zhang2015, Papadakis2016}. Each vocabulary term is represented with one numerical value in a feature vector of a document: the feature value can be calculated in different ways. The simplest is to measure the frequency of a term (\emph{tf}) in a given document. A commonly used measure is also term frequency inverse document frequency (\emph{tf-idf}), where the term frequency (\emph{tf}) is multiplied by the reciprocal frequency of the term in the entire document collection (\emph{idf}). In this way, \emph{tf-idf} reduces the importance of  the terms that appear in many documents and increased the importance of rare terms.

The major characteristic of the BOW model is the high dimensionality of the feature space: the size of the vocabulary can be tens or hundreds of thousands of terms for an average-sized document collection. Usually, to reduce the vocabulary size, the documents are first preprocessed by removing non-informative terms (stop words). Furthermore, document frequency thresholding~\citep{yang1997comparative} removes terms with document frequency below some predetermined threshold. Finally, the standard methods for feature selection or dimensionality reduction, such as principal component analysis~\citep{Jolliffe2014}, are applied. 

Traditionally, the BOW model is used as the state-of-the-art document representation model in many natural language processing applications. Its success emerges from the implementation simplicity and the fact that it often leads to high accuracy document representation. Still, it is well known that BOW is characterized with many drawbacks such as high dimensionality, sparsity, the inability to capture semantics or any dependencies between words like simple word order. Therefore new representation models in the forms of distributed word embeddings (word2vec and doc2vec) and graph-of-words (GOW) have been proposed and tested to challenge the open issues in document classification.

\subsection{Continuous-Space Models}

Continuous-space word representations capture syntactic and semantic regularities in language as constant offsets between high-dimensional vectors of words (word2vec)~\citep{Mikolov2013a}.                                                                                                                                                                                                                                                                                                                                                                                                                                                                                                                                                                                                                                                                                                                                                                                                                                                                                             The word2vec model embeds the words from a given document collection in a normed vector space, where the closeness of vectors corresponds to the word semantic similarity. In turn, the features representing a word correspond to the dimensions of the vector space. 

More specifically, word2vec employs neural networks to model the relation between a given word and its context of neighboring words in the given collection of documents. The continuous bag-of-word model (CBOW) predicts the context (neighboring words) for a given word, while the continuous skip-gram model predicts a word given the context. The neural network prediction model uses a hierarchical softmax function whose structure is a binary Huffman tree and is trained using stochastic gradient descent and back propagation algorithms. To improve the computational efficiency of model training, negative sampling is used to reduce the number of distributed context vectors considered. This type of model is also referred to as neural language models~\citep{bengio2003neural}. Recently, word2vec has been shown to be successful in many natural language processing tasks ranging from sentiment analysis~\citep{Ren2016, Rexha2016, Liang2015}, topic modeling~\citep{Bicalho2017}, through document classification~\citep{Yoshikawa2014, Lilleberg2015} and name entity recognition~\citep{Tang2014, Seok2015} to machine translation~\citep{Zou2013, Freitas2016}. 

High dimensional vectors also proved to be efficient on larger linguistic units, such as pieces of text of variable length (sentences, paragraphs or documents) resulting in paragraph2vec and doc2vec models~\citep{Mikolov2013b, Le2014}. The doc2vec models are able to predict word occurrence in the context of paragraphs or documents. Hence, doc2vec has been shown to be efficient in sentiment analysis~\citep{Le2014, Djuric2015, Sanguansat2016}, information retrieval~\citep{Le2014}, document classification~\citep{Jawahar2016}, summarization~\citep{Campr2015} and question answering~\citep{Belinkov2015}.

Recently studies of the continuous space models for document classification have been primarily focused on the exploration of one isolated aspect of the system, usually comparing different classifiers and contrasting different representation models of documents against bag-of-words as a baseline model and word2vec (rarely also doc2vec) as the suggested improvement. For example, \citet{Djuric2015} compare the \emph{tf} and \emph{tf-idf} variants of the bag-of-words model with a doc2vec model in the task of hate speech detection. They show that doc2vec model outperforms both bag-of-words variants in terms of the area under receiver operating characteristic curve obtained with a classifier based on linear regression. Similarly, \citet{Sanguansat2016} shows that the doc2vec model outperforms \emph{tf} and \emph{tf-idf} variants of the bag-of-words model in the sentiment analysis tasks in Thai and English languages, regardless of the used classifier (logistic regression, na\"{i}ve Bayes or support vector machines). \citet{Jiang2016} show the combination of the bag-of-words and the continuous space models lead to a marginal performance improvement over the alternatives in the sentiment analysis task.

In this study, we use the variants of the word2vec and doc2vec models that correspond to the alternative sizes of the feature vectors extracted from the continuous space transformation. We consider each variant as a document representation model and conduct a systematic comparison thereof with the bag-of-words and network-based models.

\subsection{Network Based Models}

The recent decade has witnessed the rise of interest in the modeling and analyzing human language with complex networks~\citep{cong2014approaching, Martincic2016}. Following this paradigm, linguistic units (words, sentences or documents) can be represented as vertices, while their relations (co-occurrence, syntax dependencies, semantic relations) as edges in a graph~\citep{Martincic2016} forming a language network
\footnote{Often, the terms of graphs and networks are interchanged depending on the field (mathematics, computer science or physics): the authors also interchangeably refer to nodes or vertices and links or edges.}.
A language (linguistic) network is formalized by a pair of sets $(V,E)$ where $V$ is the set of nodes representing the linguistic units and $E$ the set of links representing the interactions between them. The network formalization captures the structural (topological) properties of a text, which is quantified through the computation of various network properties at a different scale. On the micro scale of individual nodes, we observe the role of individual nodes in the network topology, on the mezzo scale of subnetworks, we examine the structure of communities of network nodes, and on the macro scale, the properties summarize the structural characteristics of the entire network.

When the linguistic units in the language network correspond to words, we refer to them as a graph-of-words (GOW) model for document representation. There are several advantages of using GOW, grounded in the graph and complex networks' theory. First, the model is known to be robust to input noise. Additionally, GOW significantly reduces the dimensionality of the representation space, when properties on the mezzo and macro levels are being considered. Note however, that this comes at the cost of the high computation complexity of the procedures for calculating the properties. The GOW model in its diverse variants has been applied to many natural language processing tasks, including text summarization~\citep{Antiqueira2009}, keyword extraction~\citep{Beliga2015, Beliga2016}, text genre detection~\citep{grabska2012complex, Martincic2016a} and document classification~\citep{Hassan2007, Blanko2012, Rossi2014, rousseau2015text, Malliaros2015, Papadakis2016, Nguyen2016}.

Note that the GOW model does not come with a standardized language network representation and set of features (network properties). The diversity of the network based models is related to the variety of networks, ranging from directed and undirected through unweighted and weighted to bipartite graphs. Moreover, it seems that there is no unique strategy in utilizing micro, mezzo and macro level structural properties, which contribute even more to diversification of reported models. Similarly, \citet{Malliaros2015} substitute \emph{tf} with micro measures of node centrality (degree, in- and out- degree, closeness) obtained from the language network. \citet{Jiang2010} model documents as graphs and use weighted frequencies to extract frequent subgraphs on the mezzo level, counts of which are used as features. \citet{rousseau2015text} also exploit frequent subgraphs extracted from the networks as features. In addition, they examine the main-core of the language network as a technique for the reduction of the dimensionality of feature vectors.

\citet{Hassan2007} use algorithm for random walk through the language network to measure term properties that replace the \emph{tf} metric of the bag-of-words model, which leads to the significant improvement of the performance in document classification tasks regardless of the classifier. They perform the analysis on the two benchmarks also used in our study. \citet{Blanko2012} propose a representation of documents with page ranks of nodes in a network constructed from text. Additionally, they employ macro and mezzo level measures of average page length and clustering coefficients. Beside co-occurrences they also incorporate grammatical relations (part-of-speech tags) as directed or undirected network links.

\citet{Rossi2014, Rossi2016} represent documents and classes as bipartite networks, and induce the weights on the links using the least mean square method. Induced weights are used as class model for the classification of unseen documents. Similarly, \citet{Papadakis2016} employ per-class networks constructed from character or word n-grams in a document. Classification is based on network similarity measures quantified as a Jaccard overlap of links or weighted overlap between the network of the new document and class baseline network.

\citet{Nguyen2016} represent documents with the undirected bipartite graphs as the underlying  structures of restricted Bolzman machines. They show that the application of the structure of features represented in a form of graph improves interpretability of the topics and contributes to the classification performance. 

In this study, we employ a variety of network properties at all three levels simultaneously. For the properties at the micro (node) level, we consider different methods for their aggregation into document features.

\section{Experimental Setup}

In this section, we present details of the setup of the empirical comparison of the different document representation models for document classification. First, we describe the data sets used in the experiments and the data preprocessing steps. Furthermore, we elaborate upon the used implementations and peculiar values of parameters of the document representation and classification models. Finally, we introduce the performance metrics methods used for the evaluation and ranking of the models.

\subsection{Data and Preprocessing}

Table~\ref{dataset-properties} provides an overview of the properties of the four data sets used in experiments. They represent a standard set of benchmarks for various natural language processing and text mining tasks and have been used in numerous other studies \citep{Hassan2007, Ren2013, Yogatama2014, Yoshikawa2014, Malliaros2015, rousseau2015text, Rossi2016, Papadakis2016, Uysal2016, Nguyen2016}.

\begin{table}[!ht]
\caption{\textbf{Data sets properties.}}
\begin{center}
\begin{tabular}{lrrrr}
\hline
				Property/Data set               & Brown        & 20News        & Reuters8 & WebKB   \\ \hline
				\# of documents                  & 500          & 18,846        & 9,460     & 8,274   \\
				\# of different words            & 32,174       & 173,296       & 37,074    & 103,847   \\ \hline
				\# of words (document length)    & 541,073      & 3,114,002     & 851,635   & 1,894,406 \\
				minimum document length          & 188          & 10            & 6         & 1       \\
				maximum document length          & 957          & 8,407         & 484       & 9,294    \\
				average document length          & 593          & 111           & 58        & 119     \\ \hline
				\# of labels                     & 2, 4, 10, 15 & 20            & 8         & 7       \\
				labels type                      & Genre/Topic  & Topic         & Topic     & Topic   \\
				labels hierarchy                 & Yes          & No            & No        & No      \\ \hline
\end{tabular}
\begin{flushleft}
Properties of the four benchmark data sets for document classification are reported for preprocessed documents, that is after the tokenization, removal of stop words and stemming. Document length (minimum, maximum and average) is reported as a number of word stems.
\end{flushleft}
\label{dataset-properties}
\end{center}
\end{table}

The Brown corpus consists of 500 documents of over 2,000 tokens each, which are written in a wide range of styles and a variety of prose~\citep{francis1979brown}. There are 15 document classes structured in a taxonomy consisting of four levels with 2, 4, 10, and 15 class labels, respectively. Therefore, in the experiments we consider four different document classification tasks related to the Brown corpus, referred to as Brown$n$, where $n$ represents the number of class labels (2, 4, 10 or 15). We use the version of the Brown corpus included in the Python Natural Language Toolkit~\citep{bird2009natural}.

Twenty Newsgroups or 20News corpus~\footnote{Lang K. The 20News data set. 2004. http://qwone.com/~jason/20Newsgroups/}~\citep{ lang1995newsweeder} is a set of almost 19 thousand newsgroup posts on twenty topics. In the experiments, we consider each topic to represent a document class. The corpus was taken from the Python scikit-learn library for machine learning~\citep{sklearn_api}.

Reuters8~\footnote{http://www.daviddlewis.com/resources/testcollections/reuters21578/} is a subset of the Reuters-21578 collection of news articles that includes the articles from the eight most frequent classes (acq, crude, earn, grain, interest, money-fx, ship, trade)~\citep{lewis2004rcv1}.

WebKB~\footnote{http://www.cs.cmu.edu/afs/cs/project/theo-20/www/data/}~\citep{craven1998learning} is a corpus of Web pages collected from computer science departments of four universities in January 1997. The class labels are faculty, staff, department, course, project, student and other. The Web pages are included in the corpus as HTML documents, so we have employed the Python library Beautiful Soup \footnote{https://www.crummy.com/software/BeautifulSoup/bs4/doc/index.html} to extract the text from the HTML pages.
		
The first step in natural language processing, also necessary when performing document classification, is the preprocessing of text in documents. The preprocessing typically includes document tokenization, the removal of stop words and normalization. During tokenization, the document is broken down into lexical tokens: in our case, we use words. Removing stop words is the process of removing frequently used words that are the most common, short function words which do not carry strong semantic properties, but are needed for the syntax of language (for example, pronouns, prepositions, conjunctions, abbreviations and interjections). We use the list of English stop words from the Python Natural Language Toolkit (NLTK). In the last phase of document normalization, we perform the reduction of different inflectional word forms into a single base word form. More specifically, we use stemming, a simple heuristic process of shortening the different word forms to a common root referred to as a stem. To this end, we employ the implementation of the Porter stemming heuristics~\citep{porter1980algorithm} from NLTK.

\subsection{Document Representation Models and Dimensionality Reduction}

Table~\ref{numOfFeatures} shows a number of features for the different variants of the three document representations models. The models word2vec, doc2vec and GOW retain the same number of features across the four data sets. For both variants \emph{tf} and \emph{tf-idf} of the BOW model, the number of features after PCA dimensionality reduction varies depending on the data set. For example, the dimensionality of the 20News feature space is about ten times higher than in Brown. The variants of the GOW model are always constructed from the same set of network measures and the size of features vectors in word2vec and doc2vec models is predetermined by the size of the parameter.

\begin{table}[!ht]
\caption{\textbf{Dimensionality of the feature space.}}
\begin{center}
\begin{tabular}{lrrrr}
		\hline
		Features/Data set & Brown & 20News & Reuters8 & WebKB \\ \hline
		tf+PCA            & 267   &  1,960             & 487       &  996     \\
		tf-idf+PCA        & 310   &  3,565             & 1,184      &   2,074    \\ \hline
		word2vec 25..100      & 25..100   & 25..100       &    25..100       & 25..100     \\
		doc2vec 25..1,000     & 25..1,000   & 25..1,000           & 25..1,000       &  25..1,000     \\ \hline
		GOW-average     & 19    &   19            &   19        & 19      \\
		GOW-histogram   & 128   &   128            &    128       &    128   \\
		GOW-quantiles   & 68    &   68            & 68          &  68     \\ \hline
	\end{tabular}
\begin{flushleft}
Dimensionality of the feature spaces for the variants of the three document representation models (bag-of-words, continuous space and network based) for the four benchmark data sets.
\end{flushleft}
\label{numOfFeatures}
\end{center}
\end{table}

\subsubsection{Bag of words}
Bag-of-words features are calculated with the scikit-learn library in Python~\citep{sklearn_api} using the \emph{TfidfVectorizer} function. For a bag-of-words representation of a given document $d$, we use two weighting schemas \emph{tf} and \emph{tf-idf}~\citep{manning2008h}:
\begin{itemize}
\item Term Frequency (\emph{tf}): the weight of the term $t$ in $d$ equals the number of occurrences of $t$ in $d$.  
\item Term Frequency, Inverse Document Frequency (\emph{tf-idf}): the weight of term $t$ in document $d$ equals 
${\it tf-idf}_{t,d} = {\it tf}_{t,d} \times {\it idf}_t$. The term $idf_t$ is an inverse document frequency defined as 
${\it idf}_t = log{(1+n)/(1+df_t)}+1$, where $df_t$ is the number of documents in the data set that contain $t$, and $N$ denote the total number of documents in the data set.
\end{itemize}

For calculating the features we also use document frequency thresholding~\citep{yang1997comparative, Ren2013} for removing terms with a document frequency less then 5. To further reduce the dimensionality of the feature space, we apply the principal component analysis (PCA) on the obtained feature vectors~\citep{Jolliffe2014} as implemented in the scikit-learn library. We selected the first $p$ principal components that explain at least 80\% of the total data variance (parameter value \emph{n\_componenets=0.80}) as features for document classification.
	
\subsubsection{Continuous space}
For the continuous space document representation models we use the word2vec and doc2vec methods as implemented in the gensim library~\citep{rehurek_lrec}. Word2vec implementation at input takes a list of documents, each of them being represented as a list of words, to train a neural network model, which can be used to calculate a vector representation for each word. We used the following parameter settings. The parameter \emph{min\_count} sets a lower bound of a word frequency; since we preprocessed the data set, we set this threshold to 1. The parameter \emph{size} denotes the dimensionality of the feature vectors, to this end, we use four values of 25, 50, 75 and 100. Hence, we have four variants of the word2vec model: word2vec25, word2vec50, word2vec75 and word2vec100. To get the representation of the whole document, we calculate the average of feature vectors for the words occurring in the document~\citep{Jiang2016}. For the other parameters, we retain the default settings.

The doc2vec implementation at input takes a list of documents, their unique identifiers and a list of words in each document. The trained neural network can be used to calculate a vector representation for a given document. Since the doc2vec implementation extends the word2vec class, we used the same settings of the shared parameters. In addition, we set the number of iterations over the training documents to 20, where in each iteration a random sequence of training documents is fed into the neural network. Again, we vary the dimensionality of the resulting document vectors in the interval between 25 to 1,000 leading to seven variants of the doc2vec model: doc2vec25, doc2vec50, doc2vec75, doc2vec100, doc2vec200, doc2vec500 and doc2vec1000.
	
\subsubsection{Network based}
We construct language networks with nodes representing words and links connecting adjacent words within the same sentence. The links are directed and weighted, where the weight of a link between two nodes represents the overall co-occurrence frequency of the corresponding words, while the directions represent the ordering of linguistic units in a co-occurrence pair~\citep{Martincic2016}. Although language networks are very often constructed from raw (not preprocessed text), here we apply network construction methods after tokenization, the removal of stop words and stemming. Network construction and analysis is implemented using Python NetworkX software package developed for the creation, manipulation, and study of the structure, dynamics, and functions of complex networks~\citep{schult2008exploring}.

We use the following macro level properties of the language network as features: number of links, number of nodes, average degree, average shortest path, global and local efficiency. Next, we calculate local measures on the micro level of individual nodes: in-degree and out-degree, in-strength and out-strength, in-selectivity and out-selectivity, inverse participation ratio, betweenness, closeness and page rank. Finally, two mezzo level properties of transitivity and clustering coefficient are also used as features. Definitions and explanations of all the used network properties are in Appendix A.

Since the micro and mezzo level properties are measured for individual nodes, we use three different aggregation methods to construct features vectors for the whole document. First, we take the average value of the property measured at individual nodes. Second, we take the minimal, maximal value and the three quartiles of the property values distribution. Third, we put the values in a histogram with ten equidistant intervals, and we measure the frequency of values in each interval. The three aggregation methods lead to three variants of the network-based document representation model: GOW average, GOW quartiles and GOW histograms.

\subsection{Learning and Evaluating Classification Models}

Once we have documents represented with features, we can use an arbitrary machine learning method for the supervised learning of the task of document classification. In the experiments, performed here, we use random forest~\citep{Breiman2001}, a strong and robust classification model that is also versatile; it is reported to work well in a variety of contexts, domains and data sets~\citep{bosch2007image, dubath2011random, ellis2014, Onan2016}.

To obtain an unbiased, out-of-sample estimate of the classification performance, we use a single split of the training data set into training and test data using \emph{createDataPartition} from the caret package in R~\citep{caret}. Two of the experimental data sets (20News, Reuters8) already cluster their documents into training and test sets, while for the other two, we take a random, stratified, 80\% samples of documents without repetition as a training set and the remaining 20\% of documents as a test set as presented in Table~\ref{numTrainTest}. Note that the samples were stratified with respect to the distribution of the document class labels.

\begin{table}[!ht]
\caption{\textbf{Splitting of the data sets.}}
\begin{center}
\begin{tabular}{lrrr}
\hline
			& \begin{tabular}[c]{@{}l@{}}Number of\\ train documents\end{tabular} & \begin{tabular}[c]{@{}l@{}}Number of\\ test documents\end{tabular} & Total \\ \hline
			Brown         & 401  (80\%)                                                               & 99                                                                (20\%) & 500     \\ \hline
			20News & 11,314 (60\%)                                                              & 7,532                                                 (40\%) & 18,846     \\ \hline
			Reuters8     & 6,800 (72\%)                                                               & 2,660                                                              (28\%) & 9,460     \\ \hline
			WebKB         & 6,623 (80\%)                                                               & 1,651                                                                 (20\%) & 8,274     \\ 
\hline
\end{tabular}
\begin{flushleft}
Number of train and test documents in each of the experimental data sets: 20News and Reuters8 data sets are already split into training and test sets of documents, for the other two data sets, we used stratified sampling of the data sets to obtain training and test sets.
\end{flushleft}
\label{numTrainTest}
\end{center}
\end{table}

Another reason for selecting the random forest classifier is its robustness to the different parameter settings. Following other applications of random forest, we only tune the value of the parameter {\sl mtry}, that is the number of feature candidates considered for selecting a tree split in each iteration of the tree building procedure~\citep{James:2014:ISL:2517747}. The value of {\sl mtry} parameter is tuned on the training set only using the {\sl tuneRF} function from the R package {\sl RandomForest}~\citep{randomForest} also providing the implementation of the random forest classifier, which is used in the experiments.

The commonly used measure of classification performance is accuracy, so we are going to use it for the evaluation of the models. Note however, that in document classification, we ofter encounter tasks where the distribution of class labels is highly imbalanced. Thus, accuracy does not provide sufficient insight into classification performance. To this end, we also employ the commonly used area under the receiver operating characteristic curve (AUROC). In addition, we also use three per-class measures of recall $ {TP_i} / {(TP_i + FN_i)} $, precision $ {TP_i}/{(TP_i + FP_i)} $, and F1-score $ {2 \times {\it prec} \times {\it recall}}/{({\it prec} + {\it recall})} $, where $i$ denotes the class label, while $TP_i$, $FP_i$, and $FN_i$ denote the number of true positives, false positives and false negatives for the class label $i$, respectively.
	
\subsection{Ranking Classification Models}

The ranking classification model with regard to a single performance measure, such as accuracy, is trivial: the larger the performance metrics, the better the classification model is. In contrast, when we rank models with regard to a single per-class measure (recall, precision, F1-score or AUROC), we have to compare their performance along multiple dimensions, that is for each class separately. In that case, the ranking for one class label can be different from the rankings for the other class labels. Thus, the issue of the overall ranking with regard to a single measure (for example, recall) becomes non-trivial.

To obtain a ranking along multiple dimensions, we employ a method from multi-objective decision theory~\citep{srinivas1994muiltiobjective}. First, we embed the classification models in the multidimensional space, where each dimension corresponds to a per-class performance measure for a single class: each classification model represents a single point in that space. To identify top-ranked models, we search for a set of non-dominated points in the space. These points correspond to models that are best performers according to at least one per-class dimension (in other words, we identify the Pareto front of non-dominated points). After we assign the top ranks to these models, we remove the corresponding points from the multidimensional space and recursively continue with ranking until all the models are ranked.

\section{Experimental Results}
This section provides an overview of experimental results presented in a form of rankings of classification models (and the corresponding document representation models) according to the non-dominated sorting algorithm presented above. Beside rankings on each task separately, we also present the average rankings achieved on all seven document classification tasks. We rank the 16 classification models corresponding to the 16 variants of the three document representation models: two \textbf{bag-of-words} (\emph{tf} and \emph{tf-idf }variants), four \textbf{word2vec} (the dimensionality of the feature vectors increasing from 25 to 100), seven \textbf{doc2vec} (the dimensionality of feature vectors increasing from 25 to 1,000) and three \textbf{graph-of-words} (averages, quartiles and histograms) variants.

The rankings according to the accuracy (Table~\ref{accuracyRanks}), AUROC (Table~\ref{aucRanks}) and 
F1-score (Table~\ref{f1Ranks}) are presented in the following three subsections. Per-class sores for precision and recall are in Tables~\ref{precisionRanks} and \ref{recallRanks} in Appendix B. Detailed tables with all the obtained results and absolute (not ranked) performance measures per task, class label and document representation model are available in Appendix C.  
 
In addition to the analysis of the overall rankings, we are going to analyze the rankings of a groups of data sets (or tasks), clustered according to the additional criteria, such as average document size, the vocabulary size and the number of class labels. The average document size is considered as long in Brown, medium in 20News ans WebKD and short in Reuters8 data set. According to the vocabulary size, we can group the data sets into smaller (Brown and Reuters8, cca. 30K) and larger vocabulary (20News and WebKD, more than 100K). We will also consider three groups of tasks with small (2-4 for Brown2 and Brown4), moderate (7-10 for Brown10, WebKD and Reuters8) and large (15-20 for Brown15 and 20News) numbers of class labels.

\subsection{Accuracy Rankings}

\begin{table}[!ht]
\caption{\textbf{Ranking according to accuracy.}}
\begin{center}
\begin{tabular}{lrrrrrrrr}

				\hline
				& Brown2 & Brown4 & Brown10 & Brown15 & Reuters8 & WebKB & 20News & Avg. rank \\ \hline
				tf            & \textbf{1}       & 2       & \textbf{1}        & 3        & 2         & 6     & \textbf{1}             & \textbf{2.29}    \\ 
				tf-idf        & 5       & 6.5     & 5        & 6        & \textbf{1}         & 5     & 6             & 4.93    \\ \hline
				word2vec 25   & 8.5     & 10      & 11       & 9.5      & 6         & 4     & 13            & 8.86    \\ 
				word2vec 50   & 2       & 8       & 8.5      & 7.5      & 5         & 2     & 10            & 6.14    \\ 
				word2vec 75   & 5       & 15      & 10       & 12       & 3         & \textbf{1}     & 9             & 7.86    \\ 
				word2vec 100  & 5       & 10      & 12.5     & 9.5      & 4         & 3     & 8             & 7.43    \\ \hline
				doc2vec 25    & 3       & 6.5     & 2.5      & 5        & 7         & 7     & 3             & 4.86    \\ 
				doc2vec 50    & 11.5    & 3       & 2.5      & 3        & 8         & 8     & 2             & 5.43    \\ 
				doc2vec 75    & 8.5     & \textbf{1}       & 4        & \textbf{1}        & 9         & 9     & 4             & 5.21    \\ 
				doc2vec 100   & 13      & 4       & 6        & 3        & 10        & 10    & 5             & 7.29    \\ 
				doc2vec 200   & 14.5    & 5       & 7        & 7.5      & 12        & 11    & 7             & 9.14    \\
				doc2vec 500   & 16      & 10      & 8.5      & 13       & 13        & 15    & 12            & 12.50   \\
				doc2vec 1000  & 14.5    & 16      & 14       & 14       & 11        & 16    & 11            & 13.79   \\ \hline
				GOW average   & 8.5     & 12.5    & 12.5     & 11       & 16        & 14    & 16            & 12.93   \\ 
				GOW quantiles & 11.5    & 14      & 15       & 15.5     & 15        & 13    & 14            & 14.00   \\
				GOW histogram & 8.5     & 12.5    & 16       & 15.5     & 14        & 12    & 15            & 13.36   \\ \hline
\end{tabular}
\begin{flushleft}
Rankings of the document representation models by document classification task according to accuracy. The last column reports the average rankings over all tasks. Top-ranked models in each column are in bold.
\end{flushleft}
\label{accuracyRanks}
\end{center}
\end{table}

Table~\ref{accuracyRanks} summarizes the rankings of the document representation models according to the accuracy. The average ranks according to accuracy suggest that both variants (\emph{tf} and \emph{tf-idf}) of BOW outperform the other representation models. The variants of doc2vec corresponding to low-dimensional feature vectors ($ \leq 75 $) outperform all variants of word2vec, while GOW model variants (together with high-dimensional variants of doc2vec) are ranked at the bottom. Note that the GOW average is the top-ranked variant among the variants of the GOW model.

For longer documents according to accuracy the BOW and doc2vec exhibit comparable rankings. For medium sized documents, BOW and word2vec are comparable, while for short documents BOW is clearly outperforming all other representation models.

The same descendant ordering of ranked representation models BOW, doc2vec, word2vec, GOW is preserved regardless of the size of vocabulary or the number of class labels. The only slight deviation (in which word2vec and doc2vec exchange the ranking positions) is noticed for the large-vocabulary data sets and moderate number of class labels.

To sum up, according to accuracy, two variants of the BOW model are the most advisable regardless of the vocabulary size or the number of class labels, while for long documents doc2vec can be considered as well. 

\subsection{AUROC Rankings}
	
\begin{table}[!ht]
\caption{\textbf{Ranking according to AUROC.}}
\begin{center}
\begin{tabular}{lrrrrrrrr}

				\hline
				& Brown2 & Brown4 & Brown10 & Brown15 & Reuters8 & WebKB & 20News & Avg. rank \\ \hline
				tf            & 8       & 5       & \textbf{3.5}      & 9        & 4         & \textbf{5}     & 11.5          & 6.57    \\ 
				tf-idf        & 5       & 12      & 11.5     & 9        & \textbf{1}         & \textbf{5}     & \textbf{1}             & 6.36    \\ \hline
				word2vec 25   & 12      & 12      & 11.5     & 9        & 8         & \textbf{5}     & 8             & 9.36    \\ 
				word2vec 50   & 7       & 12      & 11.5     & 9        & 4         & \textbf{5}     & 4             & 7.50    \\ 
				word2vec 75   & 6       & 12      & 11.5     & 9        & 4         & \textbf{5}     & 4             & 7.36    \\ 
				word2vec 100  & 10.5    & 12      & 11.5     & 9        & 4         & \textbf{5}     & 4             & 8.00    \\ \hline
				doc2vec 25    & 4       & 5       & \textbf{3.5}      & 9        & 4         & \textbf{5}     & 4             & 4.93    \\ 
				doc2vec 50    & \textbf{1}       & \textbf{1.5}     & \textbf{3.5}      & \textbf{1}        & 8         & \textbf{5}     & 4             & \textbf{3.43}    \\ 
				doc2vec 75    & 2       & \textbf{1.5}     & \textbf{3.5}      & 9        & 4         & \textbf{5}     & 8             & 4.71    \\ 
				doc2vec 100   & 9       & 5       & \textbf{3.5}      & 9        & 11.5      & 10.5  & 8             & 8.07    \\ 
				doc2vec 200   & 4       & 5       & \textbf{3.5}      & 9        & 11.5      & 12.5  & 11.5          & 8.14    \\
				doc2vec 500   & 14      & 12      & 11.5     & 9        & 11.5      & 12.5  & 11.5          & 11.71   \\
				doc2vec 1000  & 10.5    & 12      & 11.5     & 9        & 8         & 10.5  & 11.5          & 10.43   \\ \hline
				GOW average   & 13      & 5       & 11.5     & 9        & 16        & 15    & 15            & 12.07   \\ 
				GOW quantiles & 15      & 12      & 11.5     & 9        & 14.5      & 15    & 15            & 13.14   \\ 
				GOW histogram & 16      & 12      & 11.5     & 9        & 14.5      & 15    & 15            & 13.29   \\ \hline
\end{tabular}
\begin{flushleft}
Rankings of the document representation models by document classification task according to AUROC. The last column reports the average rankings over all tasks. Top-ranked models in each column are in bold.
\end{flushleft}
\label{aucRanks}
\end{center}
\end{table}

Table~\ref{aucRanks} summarizes the rankings of the document representation models according to the AUROC performance measure. The average ranks according to AUROC suggest that doc2vec has a performance superior to all the other representation models. Again, this is only true for the variants of doc2vec corresponding to low-dimensional feature vectors ($ \leq 75 $). The other models are ranked as follows: BOW, word2vec and GOW at the bottom. Note that, just like in the case of accuracy, the GOW average is the top-ranked variant among the variants of the GOW model.

The overall average ranking is consistent among the majority of the document classification tasks. Two notable exceptions are Reuters8 and 20News, where the \emph{tf-idf} variant of the BOW model outperforms the equally ranked doc2vec and word2vec models. Two other exceptions include WebKD, where BOW, word2vec and low-dimensional variants of doc2vec are top-ranked, and Brown10, where the \emph{tf} variant of BOW is among the top-ranked models. Note that three out of four exceptions correspond to the tasks with moderate number of class labels. Other data set properties do not seem to be related to the AUROC performance.

To sum up, according to AUROC, the doc2vec variants with low-dimensional feature vectors outperform other document representation models. For tasks with a moderate number of class labels, one can also consider the bag-of-word model, although the decision for BOW calls for the careful selection of one of its two variants.
 
\subsection{F1-Score Rankings}

\begin{table}[!ht]
\caption{\textbf{Ranking according to F1-score.}}
\begin{center}
\begin{tabular}{lrrrrrrrr}				
				\hline
				& Brown2 & Brown4 & Brown10 & Brown15 & Reuters8 & WebKB & 20News & Avg. rank\\ \hline
				tf            & \textbf{1}       & \textbf{2}       & \textbf{2.5}      & 4        & 4         & \textbf{4}     & 9.5           & \textbf{3.86}    \\ 
				tf-idf        & 5       & 13.5    & 6.5      & 11.5     & \textbf{1}         & \textbf{4}     & \textbf{1}             & 6.07    \\ \hline
				word2vec 25   & 8.5     & 7       & 12.5     & 11.5     & 7.5       & \textbf{4}     & 9.5           & 8.64    \\ 
				word2vec 50   & 2       & 7       & 12.5     & 4        & 4         & \textbf{4}     & 9.5           & 6.14    \\ 
				word2vec 75   & 5       & 13.5    & 12.5     & 11.5     & 4         & \textbf{4}     & 9.5           & 8.57    \\ 
				word2vec 100  & 5       & 7       & 12.5     & 11.5     & 4         & \textbf{4}     & 9.5           & 7.64    \\ \hline
				doc2vec 25    & 3       & 7       & \textbf{2.5}      & 4        & 4         & \textbf{4}     & 3.5           & 4.00    \\ 
				doc2vec 50    & 11.5    & \textbf{2}       & \textbf{2.5}      & 4        & 7.5       & 8     & 3.5           & 5.57    \\ 
				doc2vec 75    & 8.5     & \textbf{2}       & \textbf{2.5}      & \textbf{1}        & 9         & 9.5   & 3.5           & 5.14    \\ 
				doc2vec 100   & 13      & 7       & 6.5      & 4        & 10        & 9.5   & 3.5           & 7.64    \\ 
				doc2vec 200   & 14.5    & 7       & 6.5      & 11.5     & 14        & 12    & 9.5           & 10.71   \\
				doc2vec 500   & 16      & 13.5    & 6.5      & 11.5     & 14        & 15    & 9.5           & 12.29   \\
				doc2vec 1000  & 14.5    & 13.5    & 12.5     & 11.5     & 11        & 15    & 9.5           & 12.50   \\ \hline
				GOW average   & 8.5     & 13.5    & 12.5     & 11.5     & 14        & 12    & 15            & 12.43   \\ 
				GOW quantiles & 11.5    & 13.5    & 12.5     & 11.5     & 14        & 15    & 15            & 13.29   \\ 
				GOW histogram & 8.5     & 7       & 12.5     & 11.5     & 14        & 12    & 15            & 11.50   \\ \hline
\end{tabular}
\begin{flushleft}
Rankings of the document representation models by document classification task according to F1-score. The last column reports the average rankings over all tasks. Top-ranked models in each column are in bold.
\end{flushleft}
\label{f1Ranks}
\end{center}
\end{table}

Table~\ref{f1Ranks} summarizes the rankings of the document representation models according to the accuracy. The average ranks according to F1-score suggest that the \emph{tf} variant of BOW outperforms the other representation models. Note however, that the low-dimensional variants of doc2vec have comparable average ranks. The \emph{tf-idf} variant of BOW and word2vec exhibit lower ranks, while the high-dimensional variants of doc2vec and all variants of GOW are ranked at the bottom. This time, the GOW histogram is top-ranked among the GOW variants.

For longer documents according to the F1-score, the doc2vec and BOW exhibit comparable rankings. For medium and small sized documents, the \emph{tf-idf} variant of BOW (and word2vec in the case of WebKB) has a slight edge over doc2vec.

The same descendant ordering of ranked representation models BOW and doc2vec on top, word2vec in the middle, and GOW at the bottom, is preserved regardless of the size of the vocabulary or the number of class labels. Note that, for the small number of class labels, GOW's performance becomes comparable to some of the lower-ranked variants of doc2vec and word2vec.

To sum up, according to the F1-score, doc2vec variants with low-dimensional feature vectors and BOW outperform other document representation models. For a data set with small sized documents, one should select the \emph{tf-idf} variant of the BOW model, for the others, low-dimensional variants of doc2vec are to be preferred. The precision and recall rankings, reported in Tables~\ref{precisionRanks}~and~\ref{recallRanks}, mostly confirm the regularities observed here, but provide some further insight: word2vec model leads to high-precision classification models.

\subsection{Discussion}

Taken together, the presented results identify two top-performing document representation models: traditionally and commonly used bag-of-words, and the emerging doc2vec model. The finding is consistent regardless of the performance evaluation metrics. The standard variant of the bag-of-words model often used in the text mining studies \citep{Yang1999, sebastiani2002machine} is \emph{tf-idf}. However, our results show that the \emph{tf} variant is often better or comparable to \emph{tf-idf}: the only performance metrics where the latter outperforms \emph{tf} is the recall. When it comes to the variants of doc2vec, we have an important result, not reported in related studies. Namely, the low-dimensional variants of doc2vec are better performers than the high-dimensional one. In our results, the phase switch is observed at 75 features. While this might be an artifact of the data sets and tasks selected in this study, it is a general pattern that consistently appears among all of them.

The word2vec model is mostly ranked in the middle. The top-ranked variants of word2vec correspond to the feature vector dimensionality of 50 and 75, the other two settings of 25 and 100 underperform in almost all the experiments. Early in the experiments, we noticed that the higher-dimensional variants ($ > 100 $) of word2vec (just like the ones for doc2vec) have a deteriorated performance. Note that the lower performance of the word2vec document representation model (when compared to doc2vec) might be due to the averaging method used \citep{Jiang2016}. On the other hand, (an extension of) doc2vec has been already shown to perform well on the document classification task \citep{Jawahar2016}.

The network-based model is systematically underperforming regardless of the task or evaluation metrics. Still, some additional remarks should be noted. The number of features used in all three network-based model variants with averaging, quartiles and histograms are 19, 68 and 128, respectively, which is lower than in other models (see Table~\ref{numOfFeatures} for details). Although GOW is lagging behind in all rankings, in some occasions it can still be the representation of choice, especially if we look for a low-dimensional document representation model robust to noise. Namely, note that the network-based model requires no extensive text preprocessing, which can be useful for fast document classification implementation in low-resources languages, which lack text preprocessing tools and resources.

Finally, the analysis of the results with respect to the criteria of the document, vocabulary and class label set size does not reveal clear results with several notable exceptions, addressed next. Regarding the document size, bag-of-words is the preferred model for smaller documents regardless of the evaluation metrics, while for larger documents doc2vec has a slight advantage. Additionally, when it comes to selecting among the word2vec and doc2vec, the latter seems to be a consistently better choice, except when observing accuracy on data sets with larger vocabularies and tasks with a moderate number of class labels.

\section{Conclusion and Further Work}

In this study we conduct a comparative analysis of document representation models for the classification task. In particular, we consider the most often used family of bag-of-words models, recently proposed continuous space models word2vec and doc2vec, and the model based on the representation of text documents as language networks (graph-of-words). While the bag-of-word models have been extensively used for the document classification task, the performance of the other two models in the same task, especially the network-based model, have not been well understood. In this study, we measure the performance of the document classifiers trained using the method of random forests on features generated from the three representation models and their derivatives. The document representation models are tested on four data sets and seven tasks enabling insights into the document classification for different document and vocabulary sizes and different number of class labels. The comparative analysis is conducted through the framework based on the non-dominated sorting of points in the multidimensional space of multiple performance measures.

To conclude, the results promote the use of both standard bag-of-words and the emerging doc2vec document representation models on new document classification tasks. Moreover, we suggest comparing their performance with multiple evaluation metrics simultaneously. In general, bag-of-words is more demanding for implementation and requires the representation vectors of higher dimensionality, which are consequently reduced with some dimensionality reduction technique (a principle component analysis in this study), which are computationally expensive. The reduced dimensionality of the features vectors space is still higher than the one in doc2vec, which deteriorates the efficiency of the classifier during construction as well as during the classification stage. On the other hand, doc2vec is in general faster, generates lower-dimensional feature vectors (up to 75), while achieving the performance comparable to that of the bag-of-words model. Additionally, doc2vec enables the fast training of the classification model and requires no further dimensionality reduction. Taking into account all this desirable characteristics of doc2vec one would prefer to use it for the document classification task. Still, some drawbacks should be considered as well. Feature vectors in doc2vec carry no meaning unless projected to a low dimensional vector space. Thus, the doc2vec model is lacking understandability, which in turn complicates the development and fine tuning of the classification model.

This study sheds some light onto possible document representation models, providing an objective and systematic evaluation and comparison in the carefully designed experimental environment. The study confirms many benefits of the state-of-the-art approaches and clarifies the behavior of newly proposed models. Although we aimed at the complete study of open issues of document representation models for the document classification task we have to set some limitations. In order to keep the experimental setup as steady as possible, we limited our focus to well known data sets, at the price of the relatively small document collections. Being aware of it, in the next step we plan to extend this comparison to large document collections which better mimic the real-life magnitude of the problem. Also, in future we are planing to experiment with different dimensionality reduction techniques. Then, we are planing to use PCA and other dimensionality reduction techniques instead of aggregations in the graph-of-words model and to experiment with different classification models. Principally, our future research plans include studying the potential of metal-level combinations of document representation models.

\section*{Acknowledgements}
\begin{small}
This work has been supported in part by the University of Rijeka under the LangNet project (13.13.2.2.07). The authors acknowledge the financial support from the Slovenian Research Agency (research core funding No. P5-0093).
\end{small}

\appendix
\section*{Appendix A. Complex Network Measures}
\label{network-measures}

In this section, we define all the network measures used for the construction of the features in the document representation model based on language networks. A language network is a pair of sets $(V,E)$, where $V$ is the set of nodes (vertices) representing the linguistic units and $E$ is the set of edges (links) links representing the interactions between linguistic units. 

The average shortest path is defined as 
\begin{equation} L=\frac{1}{N(N-1)}\sum_{i \ne j} d_{ij} \end{equation}
where \begin{math}d_{ij}\end{math} is a shortest path between nodes $i$ and $j$, and $N$ is the number of nodes. 

 An efficiency measure was first defined by ~\citet{latora2001efficient, latora2003economic} where they introduced
	it as a property which quantifies how efficiently information is
	exchanging over the network
	\begin{equation}\label{eglob}
	E_{glob}(G)=\frac{1}{N(N-1)}\sum_{i\neq{j\in G}} \frac{1}{d_{ij}}.
	\end{equation}
	Local efficiency is defined as  the average efficiency of the local subgraphs:
	\begin{equation}\label{eloc}
	E_{loc}=\frac{1}{N}\sum_{i \in G}E_{glob}(G_{i}),\qquad i\notin{G_i}
	\end{equation}
	where  \begin{math}G_{i}\end{math} is the subgraph of the neighbors of \emph{i}. 
	
	Next we calculate local measures: in-degree and out-degree  of node \emph{i}, denoted by 
	\begin{math}\label{degree}k_i^{in/out}\end{math} which are the number of its ingoing and
	outgoing nearest neighbours, in-strength and out-strength \begin{math}s_i^{in/out}\end{math}
	of the node \emph{i} which are the sum of its ingoing and outgoing edge weights, average strength or selectivity as:
	\begin{equation}\label{selectivity}
	e_i^{in/out}  = \frac{s_i^{in/out}}{k_i^{in/out}}.
	\end{equation}
	 
	Inverse participation ratio \begin{equation}
		Y_i^{in/out} = \sum_{j=1}^{N}\left(\frac{a_{ij}^{in/out}}{s_{i}^{in/out}}\right)
	\end{equation} where $s_{ij}^{in/out}$ indicate the sum of the weights of the edges incident upon node $i$ and $a_{i,j}^{in/out}$ is weight of the edge between node $i$ and $j$~\citep{menichetti2014weighted}.
	
	Transitivity is defined as \begin{equation}
		T = \frac{\#triangles}{\#triads}
	\end{equation} where triads are two edges with a shared node.
	
	The clustering coefficient is a measure which defines the presence
	of loops of the order three and is defined as:
	\begin{equation}\label{clustcoef}
	C_i=\frac{e_{ij}}{k_i(k_i-1)}
	\end{equation}
	where \begin{math}e_{ij}\end{math} represents the number of pairs of
	neighbours of \emph{i} that are connected.
	
	Betweeness centrality ($c_B$) and closeness centrality ($c_C$)~\citep{brandes2001faster} are
	\begin{equation}
		c_B(v) =\sum_{s \neq v \neq t \in V} \frac{\sigma_{st}(v)}{\sigma_{st}},
	\end{equation}
	\begin{equation}
		c_C(v) = \frac{1}{\sum_{t \in V}d_G(v,t)}
	\end{equation}
	where $\sigma_{st}=\sigma_{ts}$ denotes the number of shortest paths from $s \in V$ to $t \in V$, and $\sigma_{st}(v)$ denotes the number of shortest paths from $s$ to $t$ that some $v \in V$ lies on, and $d_G(s,t)$ is the distance between nodes $s$ and $t$.
	
	Page rank~\citep{page1999pagerank} of the node is based on the eigenvector centrality measure and implements the concept of 'voting'. The Page rank score of a node $v$ is initialized to a default value and computed iteratively until convergence using the following equation:

\begin{equation}
C_{PageRank}(v) = (1-d)+d\sum_{u \in N_{in}(v)}\frac{C_{PageRank}(u)}{|N_{out}(u)|} 
\end{equation} 

where $d$ is the dumping factor set between 0 and 1 (usually 0.85).

\section*{Appendix B. Rankings according to Precision and Recall}
\label{precisionRecallresults}
\subsection*{Precision Rankings}

Table~\ref{precisionRanks} summarizes the rankings of the document representation models according to the precision.

\begin{table}[!ht]
\caption{\textbf{Ranking according to precision.}}
\begin{center}
\begin{tabular}{lrrrrrrrr}				
				\hline				
				
				\hline
				& Brown2 & Brown4 & Brown10 & Brown15 & Reuters8 & WebKB & 20News & Avg. rank \\ \hline
				tf            & \textbf{1.5}  &\textbf{2}  & \textbf{3.5}   & 10    & 4.5   & \textbf{4.5}   & \textbf{1.5} & \textbf{3.93}    \\ 
				tf-idf        & 6.5     & 12.5    & \textbf{3.5}      & 10       & \textbf{1}         & \textbf{4.5}   & \textbf{1.5}           & 5.64    \\ \hline
				word2vec 25   & 10.5    & 12.5    & 11.5     & 10       & 4.5       & \textbf{4.5}   & 8             & 8.79    \\ 
				word2vec 50   & \textbf{1.5}     & 6.5     & 11.5     & \textbf{2}        & 4.5       & \textbf{4.5}  & 8             & 5.50    \\ 
				word2vec 75   & 6.5     & 12.5    & \textbf{3.5}      & \textbf{2}       & 4.5       &\textbf{4.5}   & 8        & 5.93    \\ 
				word2vec 100  & 6.5     & 6.5     & 11.5     & 10       & 4.5       & \textbf{4.5}   & 8             & 7.36    \\ \hline
				doc2vec 25    & 3.5     & 4       & \textbf{3.5}      & 10       & 4.5       & \textbf{4.5}   & 8             & 5.43    \\ 
				doc2vec 50    & 6.5     & \textbf{2}       & 11.5     & \textbf{2}        & 8.5       & \textbf{4.5}   & 8             & 6.14    \\ 
				doc2vec 75    & 3.5     & \textbf{2}     & \textbf{3.5}      & 10       & 8.5       & 12.5  & 8             & 6.86    \\ 
				doc2vec 100   & 14.5    & 6.5     & 11.5     & 10       & 10.5      & 12.5  & 8             & 10.50   \\ 
				doc2vec 200   & 10.5    & 6.5     & 11.5     & 10       & 14        & 12.5  & 8             & 10.43   \\
				doc2vec 500   & 14.5    & 12.5    & \textbf{3.5}      & 10       & 14        & 12.5  & 8             & 10.71   \\
				doc2vec 1000  & 10.5    & 12.5    & 11.5     & 10       & 10.5      & 12.5  & 8             & 10.79   \\ \hline
				GOW average   & 14.5    & 12.5    & 11.5     & 10       & 14        & 12.5  & 15            & 12.86   \\ 
				GOW quantiles & 14.5    & 12.5    & 11.5     & 10       & 14        & 12.5  & 15            & 12.86   \\ 
				GOW histogram & 10.5    & 12.5    & 11.5     & 10       & 14        & 12.5  & 15            & 12.29   \\ \hline
\end{tabular}
\begin{flushleft}
Rankings of the document representation models by document classification task according to precision. The last column reports the average rankings over all tasks. Top-ranked models in each column are in bold.
\end{flushleft}
\label{precisionRanks}
\end{center}
\end{table}

\subsection*{Recall Rankings}

Table~\ref{recallRanks} summarizes the rankings of the document representation models according to the recall.

\begin{table}[!ht]
\caption{\textbf{Ranking according to recall.}}
\begin{center}
\begin{tabular}{lrrrrrrrr}
				\hline
				& Brown2 & Brown4 & Brown10 & Brown15 & Reuters8 & WebKB & 20News & Avg. rank \\ \hline
				tf            & \textbf{1.5}     & 6.5     & 7        & 3.5      & \textbf{5.5}       & \textbf{3}     & 8             & 5.00    \\ 
				tf-idf        & 6.5     & 2.5     & \textbf{3}        & 3.5      & \textbf{5.5}       & \textbf{3}     & \textbf{1.5}           & \textbf{3.64}    \\ \hline
				word2vec 25   & 10.5    & 13      & 12.5     & 11       & \textbf{5.5}       & 11    & 8             & 10.21   \\ 
				word2vec 50   & \textbf{1.5}     & 2.5     & 12.5     & 11       & \textbf{5.5}       & 11    & 8             & 7.43    \\ 
				word2vec 75   & 6.5     & 13      & 12.5     & 11       & \textbf{5.5}       & 11    & 8             & 9.64    \\ 
				word2vec 100  & 6.5     & 6.5     & 12.5     & 11       & \textbf{5.5}       & 11    & 8             & 8.71    \\ \hline
				doc2vec 25    & 3.5     & 13      & \textbf{3}       & 3.5      & \textbf{5.5}       & 11    & 8             & 6.79    \\ 
				doc2vec 50    & 6.5     & 6.5     & \textbf{3}        & 11       & \textbf{5.5}       & \textbf{3}     & 8             & 6.21    \\ 
				doc2vec 75    & 3.5     & 6.5     & \textbf{3}        & \textbf{1}        & \textbf{5.5}       & \textbf{3}     & 8             & 4.36    \\ 
				doc2vec 100   & 14.5    & 6.5     & \textbf{3}        & 3.5      & \textbf{5.5}       & \textbf{3}     & \textbf{1.5}           & 5.36    \\ 
				doc2vec 200   & 10.5    & \textbf{1}       & 12.5     & 11       & 13.5      & 11    & 8             & 9.64    \\
				doc2vec 500   & 14.5    & 6.5     & 7        & 11       & 13.5      & 11    & 8             & 10.21   \\
				doc2vec 1000  & 10.5    & 13      & 12.5     & 11       & 13.5      & 11    & 8             & 11.36   \\ \hline
				GOW average   & 14.5    & 13      & 7        & 11       & 13.5      & 11    & 15            & 12.14   \\ 
				GOW quantiles & 14.5    & 13      & 12.5     & 11       & 13.5      & 11    & 15            & 12.93   \\ 
				GOW histogram & 10.5    & 13      & 12.5     & 11       & 13.5      & 11    & 15            & 12.36   \\ \hline
\end{tabular}
\begin{flushleft}
Rankings of the document representation models by document classification task according to recall. The last column reports the average rankings over all tasks. Top-ranked models in each column are in bold.
\end{flushleft}
\label{recallRanks}
\end{center}
\end{table}

\section*{Appendix C. Supplementary Material}
\label{detailedResults}

Detailed results for each class of the four data sets and document representation models with different evaluation metrics (accuracy, precision, recall, F1-score and  area under ROC curve) are provided in an online appendix file.

\section*{References}


\bibliography{references}

\end{document}


\title{Appendix C for: The Influence of Feature Representation of Text\\                     on the Performance of Document Classification}
\maketitle	
\begin{center}
\Large{\textbf{Sanda  Martin\v{c}i\'c-Ip\v{s}i\'c,  Tanja  Mili\v{c}i\'c},} \\ 
	\normalsize{smarti@uniri.hr ,  tanja.milicic@student.uniri.hr \\
		University of Rijeka, Department of Informatics, Rijeka, Croatia\\}
	\Large{\textbf{Ljup\v{c}o Todorovski} } 	\\ 
	\normalsize{Ljupco.Todorovski@fu.uni-lj.si \\
		University of Ljubljana, Faculty of Administration \\
		Institute Jo\v{z}ef Stefan, Department of Knowledge Technologies, \\
		Ljubljana} Slovenia
\end{center}

	\section{Results for Reuters8 data set}
%
%
%
%

	\begin{table}[H]		
		\resizebox*{\textwidth}{!}{%
			\begin{tabular}{llllllllllll}
				\multicolumn{2}{l}{Accuracy} & \multicolumn{2}{l}{Per class Precision} & \multicolumn{2}{l}{Per class Recall} & \multicolumn{2}{l}{Per class F1} & \multicolumn{2}{l}{Per class AUC ROC} & \multicolumn{2}{l}{Per class all} \\
				tf-idf         & 1      & tf-idf             & 1             & tf               & 5.5         & tf-idf          & 1         & tf-idf          & 1          & tf             & 5.5        \\
				tf            & 2      & tf                & 4.5           & tf-idf            & 5.5         & tf             & 4         & tf             & 4          & tf-idf          & 5.5        \\
				word2vec\_75        & 3      & doc2vec\_25             & 4.5           & doc2vec\_100           & 5.5         & doc2vec\_25          & 4         & doc2vec\_25          & 4          & doc2vec\_100         & 5.5        \\
				word2vec\_100       & 4      & word2vec\_100           & 4.5           & doc2vec\_25            & 5.5         & word2vec\_100        & 4         & word2vec\_100        & 4          & doc2vec\_25          & 5.5        \\
				word2vec\_50        & 5      & word2vec\_25            & 4.5           & doc2vec\_50            & 5.5         & word2vec\_50         & 4         & word2vec\_50         & 4          & doc2vec\_50          & 5.5        \\
				word2vec\_25        & 6      & word2vec\_50            & 4.5           & doc2vec\_75            & 5.5         & word2vec\_75         & 4         & word2vec\_75         & 4          & doc2vec\_75          & 5.5        \\
				doc2vec\_25         & 7      & word2vec\_75            & 4.5           & word2vec\_100          & 5.5         & doc2vec\_50          & 7.5       & doc2vec\_1000        & 8          & word2vec\_100        & 5.5        \\
				doc2vec\_50         & 8      & doc2vec\_50             & 8.5           & word2vec\_25           & 5.5         & word2vec\_25         & 7.5       & doc2vec\_50          & 8          & word2vec\_25         & 5.5        \\
				doc2vec\_75         & 9      & doc2vec\_75             & 8.5           & word2vec\_50           & 5.5         & doc2vec\_75          & 9         & word2vec\_25         & 8          & word2vec\_50         & 5.5        \\
				doc2vec\_100        & 10     & doc2vec\_100            & 10.5          & word2vec\_75           & 5.5         & doc2vec\_100         & 10        & doc2vec\_100         & 11.5       & word2vec\_75         & 5.5        \\
				doc2vec\_1000       & 11     & doc2vec\_1000           & 10.5          & doc2vec\_1000          & 13.5        & doc2vec\_1000        & 11        & doc2vec\_200         & 11.5       & doc2vec\_1000        & 13.5       \\
				doc2vec\_200        & 12     & doc2vec\_200            & 14            & doc2vec\_200           & 13.5        & doc2vec\_200         & 14        & doc2vec\_500         & 11.5       & doc2vec\_200         & 13.5       \\
				doc2vec\_500        & 13     & doc2vec\_500            & 14            & doc2vec\_500           & 13.5        & doc2vec\_500         & 14        & doc2vec\_75          & 11.5       & doc2vec\_500         & 13.5       \\
				gow\_histogram      & 14     & gow\_avg                & 14            & gow\_avg               & 13.5        & gow\_avg             & 14        & gow\_histogram       & 14.5       & gow\_avg             & 13.5       \\
				gow\_quantiles      & 15     & gow\_histogram          & 14            & gow\_histogram         & 13.5        & gow\_histogram       & 14        & gow\_quantiles       & 14.5       & gow\_histogram       & 13.5       \\
				gow\_avg            & 16     & gow\_quantiles          & 14            & gow\_quantiles         & 13.5        & gow\_quantiles       & 14        & gow\_avg             & 16         & gow\_quantiles       & 13.5      
			\end{tabular}}
					\caption{Ranks of document representation models for Reuters8 dataset in terms of accuracy, per class precision, per class recall, per class F1 and per class AUC ROC}
		\end{table}
	
	\begin{table}[H]
					\centering
		\begin{tabular}{lllllll}

			\hline
			Features          & Class                    & Precision      & Recall & F1   & AUC ROC  & Accuracy           \\ \hline
			\multirow{8}{*}{tf} & Acq      & 0.95           & 0.94 & 0.95 & 0.99 & \multirow{8}{*}{89.43\%} \\
			& Crude    & 0.81           & 0.81 & 0.81 & 0.99 &                   \\
			& Earn     & 0.99           & 0.96 & 0.97 & 1.00 &                   \\
			& Grain    & 0.82           & 0.83 & 0.82 & 0.99 &                   \\
			& Interest & 0.47           & 0.73 & 0.57 & 0.98 &                   \\
			& Money-fx & 0.83           & 0.67 & 0.75 & 0.98 &                   \\
			& Ship     & 0.47           & 0.71 & 0.57 & 0.98 &                   \\
			& Trade    & 0.82           & 0.87 & 0.85 & 0.99 &                   \\ \hline \hline
			\multirow{8}{*}{tf-idf} & Acq      & 0.95           & 0.97 & 0.96 & 1.00 & \multirow{8}{*}{91.20\%} \\
			& Crude    & 0.89           & 0.79 & 0.84 & 0.99 &                   \\
			& Earn     & 0.99           & 0.96 & 0.98 & 1.00 &                   \\
			& Grain    & 0.87           & 0.91 & 0.89 & 1.00 &                   \\
			& Interest & 0.49           & 0.86 & 0.62 & 0.99 &                   \\
			& Money-fx & 0.93           & 0.70 & 0.80 & 0.99 &                   \\
			& Ship     & 0.44           & 0.76 & 0.56 & 0.98 &                   \\
			& Trade    & 0.87           & 0.84 & 0.85 & 1.00 &                   \\ \hline
		\end{tabular}
				\caption{Results for 8 classes of Reuters8 data set for bag-of-words document representation models}
	\end{table}

	\begin{table}[H]
		\centering	
			\begin{tabular}{ll}
				\begin{minipage}{.5\linewidth}
				\resizebox*{\textwidth}{!}{
				\begin{tabular}[t]{lllllll}
					\hline
					Features          & Class                    & Precision      & Recall & F1   & AUC ROC  & Accuracy           \\ \hline
					\multirow{8}{*}{25} & Acq      & 0.95           & 0.93 & 0.94 & 0.99 & \multirow{8}{*}{87.86\%} \\
					& Crude    & 0.58           & 0.80 & 0.67 & 0.98 &                   \\
					& Earn     & 0.98           & 0.96 & 0.97 & 1.00 &                   \\
					& Grain    & 0.81           & 0.82 & 0.81 & 0.99 &                   \\
					& Interest & 0.42           & 0.80 & 0.55 & 0.98 &                   \\
					& Money-fx & 0.84           & 0.66 & 0.74 & 0.98 &                   \\
					& Ship     & 0.64           & 0.58 & 0.61 & 0.97 &                   \\
					& Trade    & 0.79           & 0.68 & 0.73 & 0.99 &                   \\ \hline \hline
					
					\multirow{8}{*}{50} & Acq      & 0.95           & 0.93 & 0.94 & 0.99 & \multirow{8}{*}{88.35\%} \\
					& Crude    & 0.64           & 0.79 & 0.71 & 0.98 &                   \\
					& Earn     & 0.98           & 0.96 & 0.97 & 1.00 &                   \\
					& Grain    & 0.79           & 0.87 & 0.82 & 0.99 &                   \\
					& Interest & 0.49           & 0.75 & 0.59 & 0.98 &                   \\
					& Money-fx & 0.78           & 0.67 & 0.72 & 0.98 &                   \\
					& Ship     & 0.73           & 0.60 & 0.66 & 0.97 &                   \\
					& Trade    & 0.79           & 0.73 & 0.76 & 0.99 &                   \\ \hline \hline
					
					\multirow{8}{*}{75} & Acq      & 0.96           & 0.93 & 0.94 & 0.99 & \multirow{8}{*}{88.87\%} \\
					& Crude    & 0.63           & 0.83 & 0.72 & 0.98 &                   \\
					& Earn     & 0.98           & 0.97 & 0.98 & 1.00 &                   \\
					& Grain    & 0.83           & 0.83 & 0.83 & 0.99 &                   \\
					& Interest & 0.49           & 0.77 & 0.60 & 0.98 &                   \\
					& Money-fx & 0.80           & 0.68 & 0.74 & 0.98 &                   \\
					& Ship     & 0.69           & 0.60 & 0.64 & 0.97 &                   \\
					& Trade    & 0.80           & 0.75 & 0.77 & 0.99 &                   \\ \hline \hline
					
					\multirow{8}{*}{100} & Acq      & 0.95           & 0.93 & 0.94 & 0.99 & \multirow{8}{*}{88.86\%} \\
					& Crude    & 0.64           & 0.80 & 0.71 & 0.98 &                   \\
					& Earn     & 0.98           & 0.96 & 0.97 & 1.00 &                   \\
					& Grain    & 0.81           & 0.83 & 0.82 & 0.99 &                   \\
					& Interest & 0.50           & 0.79 & 0.61 & 0.98 &                   \\
					& Money-fx & 0.83           & 0.68 & 0.75 & 0.98 &                   \\
					& Ship     & 0.63           & 0.60 & 0.61 & 0.97 &                   \\
					& Trade    & 0.81           & 0.76 & 0.79 & 0.99 &                   \\ \hline
				\end{tabular}}
				\caption{Results for 8 classes of Reuters8 data set for  word2vec document representation models with feature vectors of size 25, 50, 75 and 100}
			\end{minipage} &
	
			\begin{minipage}{.5\linewidth}

				\resizebox*{\textwidth}{!}{%
				\begin{tabular}[t]{lllllll}
					\hline
					Features          & Class                    & Precision      & Recall & F1   & AUC ROC  & Accuracy           \\ \hline		
					\multirow{8}{*}{25} & Acq      & 0.91           & 0.92 & 0.91 & 0.99 & \multirow{8}{*}{86.17\%} \\
					& Crude    & 0.68           & 0.79 & 0.73 & 0.99 &                   \\
					& Earn     & 0.98           & 0.90 & 0.94 & 0.99 &                   \\
					& Grain    & 0.85           & 0.87 & 0.86 & 1.00 &                   \\
					& Interest & 0.37           & 0.77 & 0.50 & 0.97 &                   \\
					& Money-fx & 0.80           & 0.67 & 0.73 & 0.98 &                   \\
					& Ship     & 0.42           & 0.60 & 0.49 & 0.98 &                   \\
					& Trade    & 0.74           & 0.78 & 0.76 & 0.99 &                   \\ \hline \hline
					
					\multirow{8}{*}{50} & Acq      & 0.89           & 0.87 & 0.88 & 0.99 & \multirow{8}{*}{82.11\%} \\
					& Crude    & 0.54           & 0.81 & 0.65 & 0.98 &                   \\
					& Earn     & 0.97           & 0.83 & 0.90 & 0.99 &                   \\
					& Grain    & 0.81           & 0.83 & 0.82 & 0.99 &                   \\
					& Interest & 0.33           & 0.77 & 0.46 & 0.97 &                   \\
					& Money-fx & 0.64           & 0.67 & 0.66 & 0.97 &                   \\
					& Ship     & 0.34           & 0.61 & 0.43 & 0.96 &                   \\
					& Trade    & 0.68           & 0.73 & 0.70 & 0.99 &                   \\ \hline \hline
					
					\multirow{8}{*}{75} & Acq      & 0.85           & 0.80 & 0.82 & 0.98 & \multirow{8}{*}{77.33\%} \\
					& Crude    & 0.40           & 0.82 & 0.54 & 0.98 &                   \\
					& Earn     & 0.98           & 0.77 & 0.86 & 0.98 &                   \\
					& Grain    & 0.77           & 0.85 & 0.81 & 0.99 &                   \\
					& Interest & 0.14           & 0.82 & 0.24 & 0.97 &                   \\
					& Money-fx & 0.61           & 0.64 & 0.63 & 0.97 &                   \\
					& Ship     & 0.10           & 0.64 & 0.17 & 0.95 &                   \\
					& Trade    & 0.50           & 0.64 & 0.56 & 0.98 &                   \\ \hline \hline
					
					\multirow{8}{*}{100} & Acq      & 0.82           & 0.75 & 0.79 & 0.97 &                   \\
					& Crude    & 0.34           & 0.80 & 0.48 & 0.96 & \multirow{7}{*}{73.35\%} \\
					& Earn     & 0.97           & 0.72 & 0.83 & 0.97 &                   \\
					& Grain    & 0.58           & 0.79 & 0.67 & 0.97 &                   \\
					& Interest & 0.11           & 0.88 & 0.19 & 0.95 &                   \\
					& Money-fx & 0.55           & 0.64 & 0.59 & 0.96 &                   \\
					& Ship     & 0.07           & 0.60 & 0.12 & 0.96 &                   \\
					& Trade    & 0.37           & 0.70 & 0.48 & 0.97 &                   \\ \hline
				\end{tabular}}
				\caption{Results for 8 classes of Reuters8 data set for  doc2vec document representation models with feature vectors of sizes 25, 50, 75 and 100}
			\end{minipage}
		\end{tabular}	
	\end{table}

	\begin{table}[H]
		\centering
		\begin{tabular}{ll}
			\begin{minipage}{.5\linewidth}
				\resizebox*{\textwidth}{!}{
				\begin{tabular}[t]{lllllll}
					\hline
					Features          & Class                    & Precision      & Recall & F1   & AUC ROC  & Accuracy           \\ \hline		
					\multirow{8}{*}{200} & Acq      & 0.77           & 0.69 & 0.73 & 0.97 & \multirow{8}{*}{66.77\%} \\
					& Crude    & 0.10           & 0.90 & 0.17 & 0.94 &                   \\
					& Earn     & 0.98           & 0.65 & 0.78 & 0.96 &                   \\
					& Grain    & 0.51           & 0.75 & 0.61 & 0.98 &                   \\
					& Interest & 0.02           & 1.00 & 0.03 & 0.95 &                   \\
					& Money-fx & 0.26           & 0.62 & 0.36 & 0.95 &                   \\
					& Ship     & 0.00           & 0.00 & 0.00 & 0.96 &                   \\
					& Trade    & 0.15           & 0.81 & 0.25 & 0.96 &                   \\ \hline \hline
					
					\multirow{8}{*}{500} & Acq      & 0.77           & 0.68 & 0.72 & 0.97 & \multirow{8}{*}{65.04\%} \\
					& Crude    & 0.03           & 0.86 & 0.06 & 0.95 &                   \\
					& Earn     & 0.98           & 0.63 & 0.76 & 0.96 &                   \\
					& Grain    & 0.31           & 0.96 & 0.47 & 0.99 &                   \\
					& Interest & 0.03           & 0.80 & 0.06 & 0.96 &                   \\
					& Money-fx & 0.28           & 0.63 & 0.39 & 0.96 &                   \\
					& Ship     & 0.00           & 0.00 & 0.00 & 0.96 &                   \\
					& Trade    & 0.07           & 0.89 & 0.13 & 0.96 &                   \\ \hline \hline
					
					\multirow{8}{*}{1000} & Acq      & 0.82           & 0.69 & 0.75 & 0.98 & \multirow{8}{*}{68.91\%} \\
					& Crude    & 0.08           & 0.88 & 0.15 & 0.96 &                   \\
					& Earn     & 0.97           & 0.68 & 0.80 & 0.97 &                   \\
					& Grain    & 0.38           & 0.95 & 0.54 & 0.99 &                   \\
					& Interest & 0.11           & 0.78 & 0.19 & 0.96 &                   \\
					& Money-fx & 0.40           & 0.61 & 0.48 & 0.96 &                   \\
					& Ship     & 0.03           & 1.00 & 0.07 & 0.97 &                   \\
					& Trade    & 0.24           & 0.88 & 0.38 & 0.96 &                   \\ \hline
				\end{tabular}}
				\caption{Results for 8 classes of Reuters8 data set for  doc2vec document representation models with feature vectors of sizes 200, 500 and 1000}
			\end{minipage} &
			
			\begin{minipage}{.5\linewidth}
				\resizebox*{\textwidth}{!}{
						\begin{tabular}[t]{lllllll}
							\hline
							Features          & Class                    & Precision      & Recall & F1   & AUC ROC  & Accuracy           \\ \hline		
							\multirow{8}{*}{Average} & Acq      & 0.71           & 0.50 & 0.58 & 0.81 & \multirow{8}{*}{58.31\%} \\
							& Crude    & 0.05           & 0.28 & 0.09 & 0.76 &                   \\
							& Earn     & 0.88           & 0.71 & 0.79 & 0.93 &                   \\
							& Grain    & 0.13           & 0.29 & 0.18 & 0.76 &                   \\
							& Interest & 0.09           & 0.43 & 0.15 & 0.70 &                   \\
							& Money-fx & 0.14           & 0.33 & 0.20 & 0.71 &                   \\
							& Ship     & 0.02           & 0.29 & 0.04 & 0.77 &                   \\
							& Trade    & 0.14           & 0.20 & 0.16 & 0.82 &                   \\ \hline \hline
							
							\multirow{8}{*}{Histogram} & Acq      & 0.76           & 0.48 & 0.59 & 0.82 & \multirow{8}{*}{59.66\%} \\
							& Crude    & 0.04           & 0.44 & 0.07 & 0.78 &                   \\
							& Earn     & 0.90           & 0.72 & 0.80 & 0.94 &                   \\
							& Grain    & 0.05           & 0.57 & 0.10 & 0.77 &                   \\
							& Interest & 0.08           & 0.48 & 0.14 & 0.72 &                   \\
							& Money-fx & 0.15           & 0.39 & 0.22 & 0.71 &                   \\
							& Ship     & 0.01           & 0.50 & 0.02 & 0.80 &                   \\
							& Trade    & 0.11           & 0.27 & 0.16 & 0.83 &                   \\ \hline \hline
							
							\multirow{8}{*}{Quantiles} & Acq      & 0.72           & 0.48 & 0.57 & 0.81 & \multirow{8}{*}{58.50\%} \\
							& Crude    & 0.06           & 0.41 & 0.11 & 0.78 &                   \\
							& Earn     & 0.89           & 0.72 & 0.80 & 0.94 &                   \\
							& Grain    & 0.08           & 0.24 & 0.12 & 0.77 &                   \\
							& Interest & 0.08           & 0.46 & 0.14 & 0.71 &                   \\
							& Money-fx & 0.13           & 0.35 & 0.19 & 0.71 &                   \\
							& Ship     & 0.01           & 0.50 & 0.02 & 0.79 &                   \\
							& Trade    & 0.10           & 0.18 & 0.13 & 0.83 &                  \\ \hline
						\end{tabular}}
							\caption{Results for 8 classes of Reuters8 data set for graph-of-word representation models using averaging, histograms and quantiles for aggregations of local measures}
			\end{minipage}
		\end{tabular}
	\end{table}

	\pagebreak
	
	\section{Results WebKB data set}
%
%
%
%
%
	
	\begin{table}[H]	
		\resizebox*{\textwidth}{!}{%
			\begin{tabular}{llllllllllll}
				\multicolumn{2}{l}{Accuracy} & \multicolumn{2}{l}{Per class Precision} & \multicolumn{2}{l}{Per class Recall} & \multicolumn{2}{l}{Per class F1} & \multicolumn{2}{l}{Per class AUC ROC} & \multicolumn{2}{l}{Per class all} \\
				word2vec\_75        & 1      & tf                & 4.5           & tf                & 3          & tf             & 4         & tf             & 5          & tf              & 2         \\
				word2vec\_50        & 2      & tf-idf             & 4.5           & tf-idf             & 3          & tf-idf          & 4         & tf-idf          & 5          & tf-idf           & 2         \\
				word2vec\_100       & 3      & doc2vec\_25             & 4.5           & doc2vec\_100            & 3          & doc2vec\_25          & 4         & doc2vec\_25          & 5          & doc2vec\_50           & 2         \\
				word2vec\_25        & 4      & doc2vec\_50             & 4.5           & doc2vec\_50             & 3          & word2vec\_100        & 4         & doc2vec\_50          & 5          & doc2vec\_100          & 10        \\
				tf-idf         & 5      & word2vec\_100           & 4.5           & doc2vec\_75             & 3          & word2vec\_25         & 4         & doc2vec\_75          & 5          & doc2vec\_1000         & 10        \\
				tf            & 6      & word2vec\_25            & 4.5           & doc2vec\_1000           & 11         & word2vec\_50         & 4         & word2vec\_100        & 5          & doc2vec\_200          & 10        \\
				doc2vec\_25         & 7      & word2vec\_50            & 4.5           & doc2vec\_200            & 11         & word2vec\_75         & 4         & word2vec\_25         & 5          & doc2vec\_25           & 10        \\
				doc2vec\_50         & 8      & word2vec\_75            & 4.5           & doc2vec\_25             & 11         & doc2vec\_50          & 8         & word2vec\_50         & 5          & doc2vec\_500          & 10        \\
				doc2vec\_75         & 9      & doc2vec\_100            & 12.5          & doc2vec\_500            & 11         & doc2vec\_100         & 9.5       & word2vec\_75         & 5          & doc2vec\_75           & 10        \\
				doc2vec\_100        & 10     & doc2vec\_1000           & 12.5          & gow\_avg                & 11         & doc2vec\_75          & 9.5       & doc2vec\_100         & 10.5       & gow\_avg              & 10        \\
				doc2vec\_200        & 11     & doc2vec\_200            & 12.5          & gow\_histogram          & 11         & doc2vec\_200         & 12        & doc2vec\_1000        & 10.5       & gow\_histogram        & 10        \\
				gow\_histogram      & 12     & doc2vec\_500            & 12.5          & gow\_quantiles          & 11         & gow\_avg             & 12        & doc2vec\_200         & 12.5       & gow\_quantiles        & 10        \\
				gow\_quantiles      & 13     & doc2vec\_75             & 12.5          & word2vec\_100           & 11         & gow\_histogram       & 12        & doc2vec\_500         & 12.5       & word2vec\_100         & 10        \\
				gow\_avg            & 14     & gow\_avg                & 12.5          & word2vec\_25            & 11         & doc2vec\_1000        & 15        & gow\_avg             & 15         & word2vec\_25          & 10        \\
				doc2vec\_1000       & 15     & gow\_histogram          & 12.5          & word2vec\_50            & 11         & doc2vec\_500         & 15        & gow\_histogram       & 15         & word2vec\_50          & 10        \\
				doc2vec\_500        & 16     & gow\_quantiles          & 12.5          & word2vec\_75            & 11         & gow\_quantiles       & 15        & gow\_quantiles       & 15         & word2vec\_75          & 10       
			\end{tabular}}
					\caption{Ranks of document representation models for WebKB data set in terms of accuracy, per class precision, per class recall, per class F1 and per class AUC ROC}
		\end{table}

%
	\begin{table}[H]
		\centering

		\begin{tabular}{lllllll}
			\hline
			Features          & Class                    & Precision      & Recall & F1   & AUC ROC  & Accuracy           \\ \hline
			\multirow{7}{*}{tf} 	& Course     & 0.61           & 0.75 & 0.67 & 0.95 &  \multirow{7}{*}{70.44\%} \\ 
			& Department & 0.58           & 0.95 & 0.72 & 0.96 &  \\
			& Faculty    & 0.64           & 0.67 & 0.66 & 0.94 &  \\
			& Other      & 0.86           & 0.71 & 0.78 & 0.88 &  \\
			& Project    & 0.03           & 0.75 & 0.06 & 0.88 &  \\
			& Staff      & 0.00           & 0.00 & 0.00 & 0.83 &  \\
			& Student    & 0.72           & 0.67 & 0.69 & 0.92 &                     \\ \hline \hline
			\multirow{7}{*}{tf-idf}
			& Course     & 0.68           & 0.77 & 0.72 & 0.96 &   \multirow{7}{*}{70.93\%} \\
			& Department & 0.72           & 0.81 & 0.76 & 0.97 &  \\
			& Faculty    & 0.62           & 0.70 & 0.66 & 0.93 &  \\
			& Other      & 0.83           & 0.73 & 0.78 & 0.87 &  \\
			& Project    & 0.18           & 0.56 & 0.27 & 0.86 &  \\
			& Staff      & 0.00           & 0.00 & 0.00 & 0.83 &  \\
			& Student    & 0.73           & 0.65 & 0.69 & 0.91 &                   \\ \hline
		\end{tabular}
		\caption{Results for 7 classes of WebKB data set for bag-of-words document representation models}
	\end{table}
%
	\begin{table}[H]
		\centering
		\begin{tabular}{ll}
			\begin{minipage}{.5\linewidth}
				\resizebox*{\textwidth}{!}{
				\begin{tabular}{lllllll}
					\hline
					Features          & Class                    & Precision      & Recall & F1   & AUC ROC  & Accuracy           \\ \hline
					\multirow{7}{*}{25} 
					& Course     & 0.73           & 0.82 & 0.77 & 0.97 &  \multirow{7}{*}{71.65\%}\\
					& Department & 0.47           & 0.61 & 0.53 & 0.94 &  \\
					& Faculty    & 0.65           & 0.65 & 0.65 & 0.93 &  \\
					& Other      & 0.86           & 0.75 & 0.80 & 0.89 &  \\
					& Project    & 0.21           & 0.62 & 0.31 & 0.90 &  \\
					& Staff      & 0.00           & 0.00 & 0.00 & 0.80 &  \\
					& Student    & 0.67           & 0.64 & 0.65 & 0.91 &                    \\ \hline \hline
					\multirow{7}{*}{50}
					& Course     & 0.70           & 0.80 & 0.75 & 0.97 &  \multirow{7}{*}{72.32\%}\\
					& Department & 0.47           & 0.63 & 0.54 & 0.94 &  \\
					& Faculty    & 0.67           & 0.69 & 0.68 & 0.93 &  \\
					& Other      & 0.87           & 0.75 & 0.81 & 0.89 &  \\
					& Project    & 0.20           & 0.50 & 0.29 & 0.91 &  \\
					& Staff      & 0.00           & 0.00 & 0.00 & 0.86 &  \\
					& Student    & 0.69           & 0.67 & 0.68 & 0.92 &                    \\ \hline \hline
					\multirow{7}{*}{75}
					& Course     & 0.69           & 0.82 & 0.75 & 0.97 & \multirow{7}{*}{72.74\%} \\
					& Department & 0.50           & 0.62 & 0.55 & 0.94 &  \\
					& Faculty    & 0.68           & 0.72 & 0.70 & 0.94 &  \\
					& Other      & 0.87           & 0.75 & 0.80 & 0.90 &  \\
					& Project    & 0.22           & 0.58 & 0.32 & 0.92 &  \\
					& Staff      & 0.00           & 0.00 & 0.00 & 0.84 &  \\
					& Student    & 0.70           & 0.66 & 0.68 & 0.92 &                    \\ \hline \hline
					\multirow{7}{*}{100}
					& Course     & 0.70           & 0.82 & 0.75 & 0.97 &  \multirow{7}{*}{71.96\%}\\
					& Department & 0.47           & 0.55 & 0.51 & 0.94 &  \\
					& Faculty    & 0.65           & 0.69 & 0.67 & 0.94 &  \\
					& Other      & 0.86           & 0.75 & 0.80 & 0.90 &  \\
					& Project    & 0.23           & 0.52 & 0.32 & 0.91 &  \\
					& Staff      & 0.00           & 0.00 & 0.00 & 0.84 &  \\
					& Student    & 0.70           & 0.66 & 0.68 & 0.92 &    \\             \hline
				\end{tabular}}
				\caption{Results for 7 classes of WebKB data set for word2vec  document representation models with feature vectors of sizes 25, 50, 75 and 100}
			\end{minipage} &

			\begin{minipage}{.5\linewidth}
				\resizebox*{\textwidth}{!}{
					\begin{tabular}{lllllll}
						\hline
						Features          & Class                    & Precision      & Recall & F1   & AUC ROC  & Accuracy           \\ \hline
						\multirow{7}{*}{25} 
						& Course     & 0.62           & 0.80 & 0.70 & 0.97 &  \multirow{7}{*}{69.05\%}\\
						& Department & 0.44           & 0.76 & 0.56 & 0.96 &  \\
						& Faculty    & 0.61           & 0.63 & 0.62 & 0.91 &  \\
						& Other      & 0.88           & 0.68 & 0.77 & 0.85 &  \\
						& Project    & 0.18           & 0.75 & 0.29 & 0.89 &  \\
						& Staff      & 0.00           & 0.00 & 0.00 & 0.80 &  \\
						& Student    & 0.60           & 0.69 & 0.64 & 0.90 &                   \\ \hline \hline
						\multirow{7}{*}{50}
						& Course     & 0.52           & 0.84 & 0.64 & 0.97 &  \multirow{7}{*}{65.11\%}\\
						& Department & 0.22           & 0.89 & 0.36 & 0.97 &  \\
						& Faculty    & 0.50           & 0.62 & 0.56 & 0.90 &  \\
						& Other      & 0.90           & 0.63 & 0.74 & 0.84 &  \\
						& Project    & 0.01           & 1.00 & 0.02 & 0.88 &  \\
						& Staff      & 0.00           & 0.00 & 0.00 & 0.70 &  \\
						& Student    & 0.55           & 0.67 & 0.60 & 0.90 &                   \\ \hline \hline
						\multirow{7}{*}{75}
						& Course     & 0.43           & 0.83 & 0.57 & 0.96 &  \multirow{7}{*}{62.57\%}\\
						& Department & 0.11           & 0.80 & 0.20 & 0.96 &  \\
						& Faculty    & 0.48           & 0.68 & 0.56 & 0.91 &  \\
						& Other      & 0.90           & 0.59 & 0.71 & 0.83 &  \\
						& Project    & 0.01           & 1.00 & 0.02 & 0.86 &  \\
						& Staff      & 0.00           & 0.00 & 0.00 & 0.70 &  \\
						& Student    & 0.51           & 0.69 & 0.59 & 0.90 &                  \\ \hline \hline
						\multirow{7}{*}{100}
						& Course     & 0.36           & 0.85 & 0.50 & 0.95 &  \multirow{7}{*}{60.69\%}\\
						& Department & 0.14           & 0.83 & 0.24 & 0.95 &  \\
						& Faculty    & 0.41           & 0.68 & 0.51 & 0.90 &  \\
						& Other      & 0.94           & 0.56 & 0.71 & 0.82 &  \\
						& Project    & 0.00           & 0.00 & 0.00 & 0.84 &  \\
						& Staff      & 0.00           & 0.00 & 0.00 & 0.63 &  \\
						& Student    & 0.41           & 0.73 & 0.52 & 0.88 &                   \\ \hline
					\end{tabular}}
					\caption{Results for 7 classes of WebKB data set for doc2vec  document representation models with feature vectors of sizes 25, 50, 75 and 100}
				\end{minipage}
			\end{tabular}
	\end{table}
%
%
%
%
%
	\begin{table}[H]
		\centering
		\begin{tabular}{ll}
			\begin{minipage}{.5\linewidth}
				\resizebox*{\textwidth}{!}{
				\begin{tabular}[t]{lllllll}
					\hline
					Features          & Class                    & Precision      & Recall & F1   & AUC ROC  & Accuracy           \\ \hline
					\multirow{7}{*}{200} 
					& Course     & 0.10           & 0.90 & 0.18 & 0.94 &  \multirow{7}{*}{52.76\%}\\
					& Department & 0.08           & 0.75 & 0.15 & 0.95 &  \\
					& Faculty    & 0.17           & 0.83 & 0.29 & 0.88 &  \\
					& Other      & 0.97           & 0.50 & 0.66 & 0.79 &  \\
					& Project    & 0.00           & 0.00 & 0.00 & 0.83 &  \\
					& Staff      & 0.00           & 0.00 & 0.00 & 0.61 &  \\
					& Student    & 0.24           & 0.71 & 0.36 & 0.86 &                   \\ \hline \hline
					\multirow{7}{*}{500}
					& Course     & 0.02           & 1.00 & 0.04 & 0.92 &  \multirow{7}{*}{49.55\%}\\
					& Department & 0.03           & 1.00 & 0.05 & 0.90 &  \\
					& Faculty    & 0.09           & 0.91 & 0.16 & 0.89 &  \\
					& Other      & 0.99           & 0.48 & 0.65 & 0.78 &  \\
					& Project    & 0.00           & 0.00 & 0.00 & 0.79 &  \\
					& Staff      & 0.00           & 0.00 & 0.00 & 0.64 &  \\
					& Student    & 0.15           & 0.67 & 0.25 & 0.85 &                  \\ \hline \hline
					\multirow{7}{*}{1000}
					& Course     & 0.04           & 1.00 & 0.07 & 0.94 &  \multirow{7}{*}{50.15\%}\\
					& Department & 0.03           & 1.00 & 0.05 & 0.93 &  \\
					& Faculty    & 0.08           & 0.90 & 0.16 & 0.89 &  \\
					& Other      & 1.00           & 0.48 & 0.65 & 0.80 &  \\
					& Project    & 0.00           & 0.00 & 0.00 & 0.80 &  \\
					& Staff      & 0.00           & 0.00 & 0.00 & 0.66 &  \\
					& Student    & 0.16           & 0.78 & 0.26 & 0.87 &                  \\ \hline
				\end{tabular}}
				\caption{Results for 7 classes of WebKB data set for doc2vec  document representation models with feature vectors of sizes 200, 500 and 1000}
			\end{minipage} &

			\begin{minipage}{.5\linewidth}
				\resizebox*{\textwidth}{!}{
					\begin{tabular}[t]{lllllll}
						\hline
						Features          & Class                    & Precision      & Recall & F1   & AUC ROC  & Accuracy           \\ \hline
						\multirow{7}{*}{Average} 
						& Course     & 0.06           & 0.41 & 0.10 & 0.68 &  \multirow{7}{*}{50.39\%}\\
						& Department & 0.14           & 0.83 & 0.24 & 0.71 &  \\
						& Faculty    & 0.15           & 0.29 & 0.20 & 0.72 &  \\
						& Other      & 0.81           & 0.56 & 0.66 & 0.73 &  \\
						& Project    & 0.00           & 0.00 & 0.00 & 0.68 &  \\
						& Staff      & 0.00           & 0.00 & 0.00 & 0.56 &  \\
						& Student    & 0.54           & 0.43 & 0.48 & 0.80 &                    \\ \hline \hline
						\multirow{7}{*}{Histogram}
						& Course     & 0.06           & 0.61 & 0.11 & 0.70 &  \multirow{7}{*}{51.48\%}\\
						& Department & 0.14           & 0.83 & 0.24 & 0.76 &  \\
						& Faculty    & 0.15           & 0.46 & 0.22 & 0.74 &  \\
						& Other      & 0.84           & 0.54 & 0.66 & 0.72 &  \\
						& Project    & 0.00           & 0.00 & 0.00 & 0.70 &  \\
						& Staff      & 0.00           & 0.00 & 0.00 & 0.57 &  \\
						& Student    & 0.52           & 0.44 & 0.47 & 0.80 &                   \\ \hline \hline
						\multirow{7}{*}{Quantiles}
						& Course     & 0.02           & 0.43 & 0.03 & 0.69 &  \multirow{7}{*}{50.58\%}\\
						& Department & 0.11           & 0.80 & 0.20 & 0.73 &  \\
						& Faculty    & 0.10           & 0.47 & 0.16 & 0.71 &  \\
						& Other      & 0.81           & 0.54 & 0.65 & 0.73 &  \\
						& Project    & 0.00           & 0.00 & 0.00 & 0.71 &  \\
						& Staff      & 0.00           & 0.00 & 0.00 & 0.53 &  \\
						& Student    & 0.59           & 0.42 & 0.49 & 0.80 &                   \\ \hline
					\end{tabular}}
					\caption{Results for 7 classes of WebKB data set for graph-of-word representation models using averaging, histograms and quantiles for aggregations of local measures}
				\end{minipage}
			\end{tabular}				
	\end{table}
	
	\section{Results for 20Newsgroups data set}
%
	\begin{table}[H]	
		\resizebox*{\textwidth}{!}{%
\begin{tabular}{llllllllllll}
			\multicolumn{2}{l}{Accuracy} & \multicolumn{2}{l}{Per class Precision} & \multicolumn{2}{l}{Per class Recall} & \multicolumn{2}{l}{Per class F1} & \multicolumn{2}{l}{Per class AUC ROC} & \multicolumn{2}{l}{Per class all} \\
			tf-idf         & 1      & tf-idf              & 1.5          & tf-idf            & 1.5         & tf-idf          & 1         & tf-idf          & 1          & tf-idf           & 1         \\
			doc2vec\_50         & 2      & doc2vec\_50              & 1.5          & doc2vec\_100           & 1.5         & doc2vec\_100         & 3.5       & doc2vec\_25          & 4          & tf              & 7.5       \\
			doc2vec\_25         & 3      & tf                 & 8            & tf               & 8           & doc2vec\_25          & 3.5       & doc2vec\_50          & 4          & doc2vec\_100          & 7.5       \\
			doc2vec\_75         & 4      & doc2vec\_100             & 8            & doc2vec\_1000          & 8           & doc2vec\_50          & 3.5       & word2vec\_100        & 4          & doc2vec\_1000         & 7.5       \\
			doc2vec\_100        & 5      & doc2vec\_1000            & 8            & doc2vec\_200           & 8           & doc2vec\_75          & 3.5       & word2vec\_50         & 4          & doc2vec\_200          & 7.5       \\
			tf            & 6      & doc2vec\_200             & 8            & doc2vec\_25            & 8           & tf             & 9.5       & word2vec\_75         & 4          & doc2vec\_25           & 7.5       \\
			doc2vec\_200        & 7      & doc2vec\_25              & 8            & doc2vec\_50            & 8           & doc2vec\_1000        & 9.5       & doc2vec\_100         & 8          & doc2vec\_50           & 7.5       \\
\end{tabular}}
				\caption{Ranks of document representation models for 20News data set in terms of accuracy, per class precision, per class recall, per class F1 and per class AUC ROC}
	\end{table}
	
%
%
	\begin{table}[H]
	\centering
	%
}
	\caption{Results for 20 classes of 20News data set for graph-of-word representation models using averaging, histograms and quantiles for aggregations of local measures}
	\end{minipage}
	\end{tabular}
	\end{table}
	
	\pagebreak
	\section{The results for Brown data set}
	\subsection{The results for Brown2: 2 classes}

%
%
%
%
%
%

		\begin{table}[H]
			\centering
			\resizebox*{\textwidth}{!}{%
\begin{tabular}{llllllllllll}
				\multicolumn{2}{l}{Accuracy} & \multicolumn{2}{l}{Per class Precision} & \multicolumn{2}{l}{Per class Recall} & \multicolumn{2}{l}{Per class F1} & \multicolumn{2}{l}{Per class AUC ROC} & \multicolumn{2}{l}{Per class all} \\
				tf           & 1       & tf                & 1.5           & tf               & 1.5         & tf             & 1         & doc2vec\_50          & 1          & doc2vec\_25          & 1          \\
				word2vec\_50       & 2       & word2vec\_50            & 1.5           & word2vec\_50           & 1.5         & word2vec\_50         & 2         & doc2vec\_75          & 2          & tf             & 4          \\
				doc2vec\_25        & 3       & doc2vec\_25             & 3.5           & doc2vec\_25            & 3.5         & doc2vec\_25          & 3         & doc2vec\_200         & 3          & doc2vec\_50          & 4          \\
				tf-idf        & 5       & doc2vec\_75             & 3.5           & doc2vec\_75            & 3.5         & tf-idf          & 5         & doc2vec\_25          & 4          & doc2vec\_75          & 4          \\
				word2vec\_100      & 5       & tf-idf             & 6.5           & tf-idf            & 6.5         & word2vec\_100        & 5         & tf-idf          & 5          & word2vec\_50         & 4          \\
				word2vec\_75       & 5       & doc2vec\_50             & 6.5           & doc2vec\_50            & 6.5         & word2vec\_75         & 5         & word2vec\_75         & 6          & word2vec\_75         & 4          \\
				doc2vec\_75        & 8.5     & word2vec\_100           & 6.5           & word2vec\_100          & 6.5         & doc2vec\_75          & 8.5       & word2vec\_50         & 7          & tf-idf          & 8          \\
\end{tabular}}
			\caption{Ranks of document representation models for Brown2 dataset in terms of accuracy, per class precision, per class recall, per class F1 and per class AUC ROC}
		\end{table}
		
	\begin{table}[H]
		\centering
		\begin{tabular}{lllllll}
			\hline
			Features         & Class         & Precision      & Recall & F1   & AUC ROC & Accuracy          \\ \hline
			\multirow{2}{*}{tf} 
			& Fiction       & 0.92           & 0.96   & 0.94 & 0.99     & \multirow{2}{*}{96.97\%} \\
			& Informative   & 0.99           & 0.97   & 0.98 & 0.99     &                   \\ \hline \hline
			\multirow{2}{*}{tf-idf}
			& Fiction       & 0.80           & 0.95   & 0.87 & 0.99     & \multirow{2}{*}{93.94\%} \\            
			& Informative   & 0.99           & 0.94   & 0.96 & 0.99     &                   \\ \hline
		\end{tabular}
		\caption{Results for 2 classes of Brown2 data set for bag-of-words document representation models}
	\end{table}

	\begin{table}[H]
		\centering
		%
					\end{table}

		\subsection{The results for Brown4: 4 classes}
		\begin{table}[H]
			\centering
			\label{my-label2}
			\resizebox*{\textwidth}{!}{%
			\begin{tabular}{llllllllllll}
				\multicolumn{2}{l}{Accuracy} & \multicolumn{2}{l}{Per class Precision} & \multicolumn{2}{l}{Per class Recall} & \multicolumn{2}{l}{Per class F1} & \multicolumn{2}{l}{Per class AUC ROC} & \multicolumn{2}{l}{Per class all} \\
				doc2vec\_75        & 1       & tf                & 2             & doc2vec\_200           & 1           & tf             & 2         & doc2vec\_50           & 1.5       & doc2vec\_100          & 1.5       \\
				tf           & 2       & doc2vec\_50             & 2             & tf-idf            & 2.5         & doc2vec\_50          & 2         & doc2vec\_75           & 1.5       & doc2vec\_75           & 1.5       \\
				doc2vec\_50        & 3       & doc2vec\_75             & 2             & word2vec\_50           & 2.5         & doc2vec\_75          & 2         & tf              & 5         & tf              & 9.5       \\
				doc2vec\_100       & 4       & doc2vec\_25             & 4             & tf               & 6.5         & doc2vec\_100         & 7         & doc2vec\_100          & 5         & tf-idf           & 9.5       \\
				doc2vec\_200       & 5       & doc2vec\_100            & 6.5           & doc2vec\_100           & 6.5         & doc2vec\_200         & 7         & doc2vec\_200          & 5         & doc2vec\_1000         & 9.5       \\
				tf-idf        & 6.5     & doc2vec\_200            & 6.5           & doc2vec\_50            & 6.5         & doc2vec\_25          & 7         & doc2vec\_25           & 5         & doc2vec\_200          & 9.5       \\
				doc2vec\_25        & 6.5     & word2vec\_100           & 6.5           & doc2vec\_500           & 6.5         & gow\_histogram       & 7         & gow\_avg              & 5         & doc2vec\_25           & 9.5       \\
\end{tabular}}
				\caption{Ranks of document representation models for Brown4 dataset in terms of accuracy, per class precision, per class recall, per class F1 and per class AUC ROC}
		\end{table}

\begin{table}[H]
	\centering
	\begin{tabular}{lllllll}  
		\hline           
		Features         & Class         & Precision      & Recall & F1   & AUC ROC &   Accuracy                \\ \hline
		\multirow{4}{*}{tf} 
		& Fiction       & 0.96           & 0.89   & 0.92 & 0.99     &      \multirow{4}{*}{82.83\%}             \\
		& Non Fiction   & 0.77           & 0.85   & 0.81 & 0.91     &                   \\
		& Other         & 0.80           & 0.78   & 0.79 & 0.89     &                   \\
		& Tisak         & 0.76           & 0.81   & 0.79 & 0.96     &                   \\ \hline \hline
		\multirow{4}{*}{tf-idf}
		& Fiction       & 0.80           & 0.91   & 0.85 & 0.99     &     \multirow{4}{*}{75.76\%} \\
		& Non Fiction   & 0.64           & 0.88   & 0.74 & 0.89     &                   \\
		& Other         & 0.86           & 0.64   & 0.73 & 0.87     &                   \\
		& Tisak         & 0.65           & 0.79   & 0.71 & 0.93     &                   \\ \hline
	\end{tabular}
	\caption{Results for 4 classes of Brown4 data set for bag-of-words document representation models}
\end{table}

\begin{table}[H]
	\centering
	%
				\end{table}

		\subsection{The results for Brown10: 10 classes}
		\begin{table}[H]
			\centering
			\resizebox*{\textwidth}{!}{%
			\begin{tabular}{llllllllllll}
				\multicolumn{2}{l}{Accuracy} & \multicolumn{2}{l}{Per class Precision} & \multicolumn{2}{l}{Per class Recall} & \multicolumn{2}{l}{Per class F1} & \multicolumn{2}{l}{Per class AUC ROC} & \multicolumn{2}{l}{Per class all} \\
				tf           & 1       & tf                & 3.5           & tf-idf            & 3           & tf             & 2.5       & tf             & 3.5        & tf              & 2         \\
				doc2vec\_25        & 2.5     & tf-idf             & 3.5           & doc2vec\_100           & 3           & doc2vec\_25          & 2.5       & doc2vec\_100         & 3.5        & doc2vec\_25           & 2         \\
				doc2vec\_50        & 2.5     & doc2vec\_25             & 3.5           & doc2vec\_25            & 3           & doc2vec\_50          & 2.5       & doc2vec\_200         & 3.5        & doc2vec\_75           & 2         \\
				doc2vec\_75        & 4       & doc2vec\_500            & 3.5           & doc2vec\_50            & 3           & doc2vec\_75          & 2.5       & doc2vec\_25          & 3.5        & tf-idf           & 10        \\
				tf-idf        & 5       & doc2vec\_75             & 3.5           & doc2vec\_75            & 3           & tf-idf          & 6.5       & doc2vec\_50          & 3.5        & doc2vec\_100          & 10        \\
				doc2vec\_100       & 6       & word2vec\_75            & 3.5           & tf               & 7           & doc2vec\_100         & 6.5       & doc2vec\_75          & 3.5        & doc2vec\_1000         & 10        \\
				doc2vec\_200       & 7       & doc2vec\_100            & 11.5          & doc2vec\_500           & 7           & doc2vec\_200         & 6.5       & tf-idf          & 11.5       & doc2vec\_200          & 10        \\
\end{tabular}}
			\caption{Ranks of document representation models for Brown10 dataset in terms of accuracy, per class precision, per class recall, per class F1 and per class AUC ROC}
		\end{table}
			
			\begin{table}[H]
				\centering
				\begin{tabular}{lllllll}
					\hline
					Features                &   Classes     & Precision      & Recall & F1   & AUC ROC &  Accuracy            \\ \hline
					\multirow{10}{*}{tf}
					& Belle Letters & 0.73           & 0.42   & 0.54 & 0.83     &                 \multirow{10}{*}{63.92\%}  \\
					& Editorial     & 0.60           & 0.75   & 0.67 & 0.98     &                   \\
					& Fiction       & 0.96           & 0.80   & 0.87 & 0.98     &                   \\
					& Government    & 0.50           & 0.75   & 0.60 & 0.92     &                   \\
					& Hobbies       & 0.43           & 0.60   & 0.50 & 0.97     &                   \\
					& Learned       & 0.63           & 0.59   & 0.61 & 0.81     &                   \\
					& Love          & 0.11           & 0.50   & 0.18 & 0.74     &                   \\
\hline
				\end{tabular}
				\caption{Results for 10 classes of Brown10 data set for bag-of-words document representation models}
			\end{table}

\begin{table}[H]
	\centering
	\centering
	%
				\end{table}

		\subsection{The results for Brown15: 15 classes}
		\begin{table}[H]
			\centering
			\resizebox*{\textwidth}{!}{%
			\begin{tabular}{llllllllllll}
				\multicolumn{2}{l}{Accuracy} & \multicolumn{2}{l}{Per class Precision} & \multicolumn{2}{l}{Per class Recall} & \multicolumn{2}{l}{Per class F1} & \multicolumn{2}{l}{Per class AUC ROC} & \multicolumn{2}{l}{Per class all} \\
				doc2vec\_75        & 1       & doc2vec\_50              & 2            & doc2vec\_75            & 1           & doc2vec\_75          & 1         & doc2vec\_50            & 1        & tf              & 8.5       \\
				tf           & 3       & doc2vec\_75              & 2            & tf               & 3.5         & tf             & 4         & tf               & 9        & tf-idf           & 8.5       \\
				doc2vec\_100       & 3       & word2vec\_50             & 2            & tf-idf            & 3.5         & doc2vec\_100         & 4         & tf-idf            & 9        & doc2vec\_100          & 8.5       \\
				doc2vec\_50        & 3       & tf                 & 10           & doc2vec\_100           & 3.5         & doc2vec\_25          & 4         & doc2vec\_100           & 9        & doc2vec\_1000         & 8.5       \\
				doc2vec\_25        & 5       & tf-idf              & 10           & doc2vec\_25            & 3.5         & doc2vec\_50          & 4         & doc2vec\_1000          & 9        & doc2vec\_200          & 8.5       \\
				tf-idf        & 6       & doc2vec\_100             & 10           & doc2vec\_1000          & 11          & word2vec\_50         & 4         & doc2vec\_200           & 9        & doc2vec\_25           & 8.5       \\
				doc2vec\_200       & 7.5     & doc2vec\_1000            & 10           & doc2vec\_200           & 11          & tf-idf          & 11.5      & doc2vec\_25            & 9        & doc2vec\_50           & 8.5       \\
\end{tabular}}
			\caption{Ranks of document representation models for Brown15 dataset in terms of accuracy, per class precision, per class recall, per class F1 and per class AUC ROC}
		\end{table}

	\begin{table}[H]
		\centering
		\begin{tabular}{lllllll}
			\hline
			Features          & Classes          & Precision      & Recall & F1   & AUC ROC &      Accuracy             \\ \hline
			\multirow{15}{*}{tf}
			& Adventure     & 0.60           & 0.43   & 0.50 & 0.96     &                   \multirow{15}{*}{46.24\%}  \\
			& Belle Letters & 0.80           & 0.46   & 0.59 & 0.87     &                   \\
			& Editorial     & 0.80           & 0.67   & 0.73 & 0.93     &                   \\
			& Fiction       & 0.20           & 0.17   & 0.18 & 0.90     &                   \\
			& Government    & 0.33           & 0.50   & 0.40 & 0.83     &                   \\
			& Hobbies       & 0.43           & 0.60   & 0.50 & 0.92     &                   \\
			& Humor         & 0.00           & 0.00   & 0.00 & 0.97      &                   \\
\hline
		\end{tabular}
		\caption{Results for 15 classes of Brown15 data set for bag-of-words document representation models}
	\end{table}

	\begin{table}[H]
		\centering
		%
}
		\caption{Results for 15 classes of Brown15 data set for graph-of-word representation models using averaging, histograms and quantiles for aggregations of local measures}
	\end{minipage}
	\end{tabular}
	\end{table}